%% file: main.tex
\documentclass[sigconf, preprint]{acmart}

\usepackage{amsmath}
\DeclareMathOperator*{\argmax}{\mathbf{argmax}}
\usepackage{algorithm}
\usepackage{algorithmic}
\usepackage{subfig}
\usepackage{wrapfig}
\usepackage{lipsum}
\usepackage{multicol}

\AtBeginDocument{%
  \providecommand\BibTeX{{%
    \normalfont B\kern-0.5em{\scshape i\kern-0.25em b}\kern-0.8em\TeX}}}

\providecommand{\mypara}[1]{\smallskip\noindent\emph{#1}}
\providecommand{\myparab}[1]{\smallskip\noindent\textbf{#1}}

\usepackage{mathtools}

\newcommand{\reals}{\mathbf{R}}

\newcommand{\cald}{\mathcal{D}}

\newcommand{\cov}{\mathrm{\cov}}

\newcommand{\bD}{\mathbb{D}}

\newcommand{\dkl}{\mathrm{D_{kl}}}

\usepackage{enumitem}

\setcopyright{none}
\renewcommand\footnotetextcopyrightpermission[1]{} 

\begin{document}

\title{On Extending NLP Techniques from the Categorical to the Latent Space: KL Divergence, Zipf’s Law, and Similarity Search}



\author{Adam Hare}
\orcid{0000-0003-3681-6135}
\affiliation{%
  \institution{Zoomi, Inc}
  \streetaddress{640 Lee Road}
  \city{Wayne}
  \state{Pennsylvania}
  \country{USA}
  \postcode{19087--5600}
}
\email{adam.hare@zoomi.ai}
\authornote{Corresponding authors}

\author{Yu Chen}
\affiliation{%
  \institution{College of William \& Mary}
  \streetaddress{251 Jamestown Rd}
  \city{Williamsburg}
  \state{Virginia}
  \country{USA}
  \postcode{23186-0003}}
\email{ychen39@email.wm.edu}
\authornotemark[1]

\author{Yinan Liu}
\affiliation{%
  \institution{Peking University}
  \city{Beijing}
  \country{China}}
\email{yinanliu@pku.edu.cn}

\author{Zhenming Liu}
\affiliation{%
  \institution{College of William \& Mary}
  \streetaddress{251 Jamestown Rd}
  \city{Williamsburg}
  \state{Virginia}
  \country{USA}
  \postcode{23186-0003}}
\email{zliu@cs.wm.edu}

\author{Christopher G. Brinton}
\affiliation{%
  \institution{Purdue University}
  \streetaddress{465 Northwestern Ave.}
  \city{West Lafayette}
  \state{Indiana}
  \country{USA}
  \postcode{47907-2040}}
\email{cgb@purdue.edu}

\begin{abstract}
\input{abstract.tex}
\end{abstract}

\maketitle
\pagestyle{plain} 
\input{intro_zliu}

\input{prelim.tex}
\input{methods.tex}
\input{experiments.tex}
\input{conclusion.tex}

\newpage



\bibliographystyle{ACM-Reference-Format}
\bibliography{references}

\input{appendix}
\end{document}

%% file: abstract.tex
Despite the recent successes of deep learning in natural language processing (NLP), there remains widespread usage of and demand for techniques that do not rely on machine learning. The advantage of these techniques is their interpretability and low cost when compared to frequently opaque and expensive machine learning models. Although they may not be be as performant in all cases, they are often sufficient for common and relatively simple problems. In this paper, we aim to modernize these older methods while retaining their advantages by extending approaches from categorical or bag-of-words representations to word embeddings representations in the latent space. First, we show that entropy and Kullback-Leibler divergence can be efficiently estimated using word embeddings and use this estimation to compare text across several categories. Next, we recast the heavy-tailed distribution known as Zipf's law that is frequently observed in the categorical space to the latent space. Finally, we look to improve the Jaccard similarity measure for sentence suggestion by introducing a new method of identifying similar sentences based on the set cover problem. We compare the performance of this algorithm against several baselines including Word Mover's Distance and the Levenshtein distance.

%% file: intro_zliu.tex
\section{Introduction}
\label{sec:intro}
Deep learning (DL) technologies developed in the recent decade have substantially enriched the means for natural language processing (NLP). Powerful models such as BERT \cite{bert}, ELMo \cite{elmo}, and GPT \cite{gpt} have demonstrated impressive predictive power and ability to understand text. Nevertheless, these models also have glaring drawbacks: it is excessively costly (in both computation and labor) to train customized models for special purpose applications (e.g., indexing internal documents in enterprise search engines), and they require sophisticated computational infrastructure for deployments. In addition, embeddings generated by these models usually map an entire sentence or document into latent positions or rely on contextual representations, so 
it is difficult to use these models to perform simple tasks such as estimating entropy of documents or studying fine-grained statistical patterns of the text (e.g., distributions at word level). These limitations prevent the wide deployment of large DL models. For example, open source search engines (Elasticsearch), documentation and bug tracking systems (Confluence \& JIRA), and apps (Line) developed by budget-constrained startups do not deploy any DL technologies despite the fact that many of their services primarily deal with text and languages. Instead, these applications exclusively rely on NLP technologies based on word frequencies that predate the deep learning era. 

Roughly speaking, word frequency-based models assume each sentence or document is associated with a vector $v \in \reals^m$, where $m$ is the size of the vocabulary set such that $v_i$ represents the number of occurrences of the $i$-th word. These models have been extensively studied and exhibit characteristics that are in sharp contrast to DL models: they have low training costs, are capable of processing a variety of NLP tasks (including those mentioned above such as word-level statistical analysis), are interpretable, and require negligible computing resources especially with the help of carefully-polished algorithmic techniques (e.g., minwise hash functions \cite{minhash}). Meanwhile, without attempting to extract deep representations of the texts or relying on transfer learning, they are much weaker in understanding syntactic and semantic relationships (e.g., unlike more advanced models, these usually cannot tell that ``king'' and ``queen'' are closely related \cite{word2vec}). As a result, these models have become increasingly out-of-fashion in NLP research.

This work aims to identify a middle ground that enables us to simultaneously benefit from the simplicity of frequency-based models and from the power of deep representations (see Fig.~\ref{fig:overview}). We use word embeddings derived from neural network representations (e.g., word2vec \cite{word2vec}, fastText \cite{fasttext}) to replace the bag-of-words representation, and ``remaster'' a rich set of algorithmic techniques and statistical analyses developed for word-frequency models. A central challenge here is that word-frequency models assume that the random variables (roughly speaking, a word corresponds to a random variable) are discrete whereas deep representations of words move these random variables to a continuous space. We need to understand how techniques optimized for discrete distributions can be generalized to continuous ones. Our remastered toolkits touch upon three important aspects of NLP:

\myparab{(1) Algorithmic aspect: divergence estimation.} We integrate a recently developed statistically sound estimator produced from the machine learning community with near-neighbor search algorithms developed from the system community to develop the first practically scalable solution to provably estimate the KL divergence between documents under the embedding model. 

\myparab{(2) Scientific aspect: heavy tail in the latent space.} An important observation of the word-frequency model is that the distributions of words have heavy tails, i.e., they follow power law distribution. We aim to understand whether the word distributions under embedding models also exhibit heavy tail property. Our idea is that in the latent space there are likely to be a few dense regions with many words clustered together and many sparse regions with few words. To study this, we first examine word embeddings to see if they capture part-of-speech (POS) data, which has been observed to be a possible explanation for Zipf's law in text~\cite{zipf_latent}. We then perform clustering on embeddings from different sources and examine cluster size for Zipfian patterns. 

\myparab{(3) Application aspect: combinatorial techniques for finding similar documents.} We propose a novel technique based on set cover problems that generalizes the Jaccard similarity to incorporate word embeddings. 
Our new similar sentence finding algorithm is better at identifying semantically related sentences that have low word overlaps with the query sentence, and thus is suitable for many interactive tasks such as creative writing in the educational setting or finding related bugs in issue tracking system. 

The purpose of this work is to perform a comprehensive ``remastering'' of low cost word based NLP tools. Thus we prioritize the breadth of the examination (i.e., touching multiple aspects of these models), sometimes at the cost of simplifying the experiments.\footnote{Code for experiments will be made available at \url{https://github.com/ahare63/categorical-to-latent}}

%% file: prelim.tex
\begin{figure}[!t]
\centering
\includegraphics[width=1\linewidth]{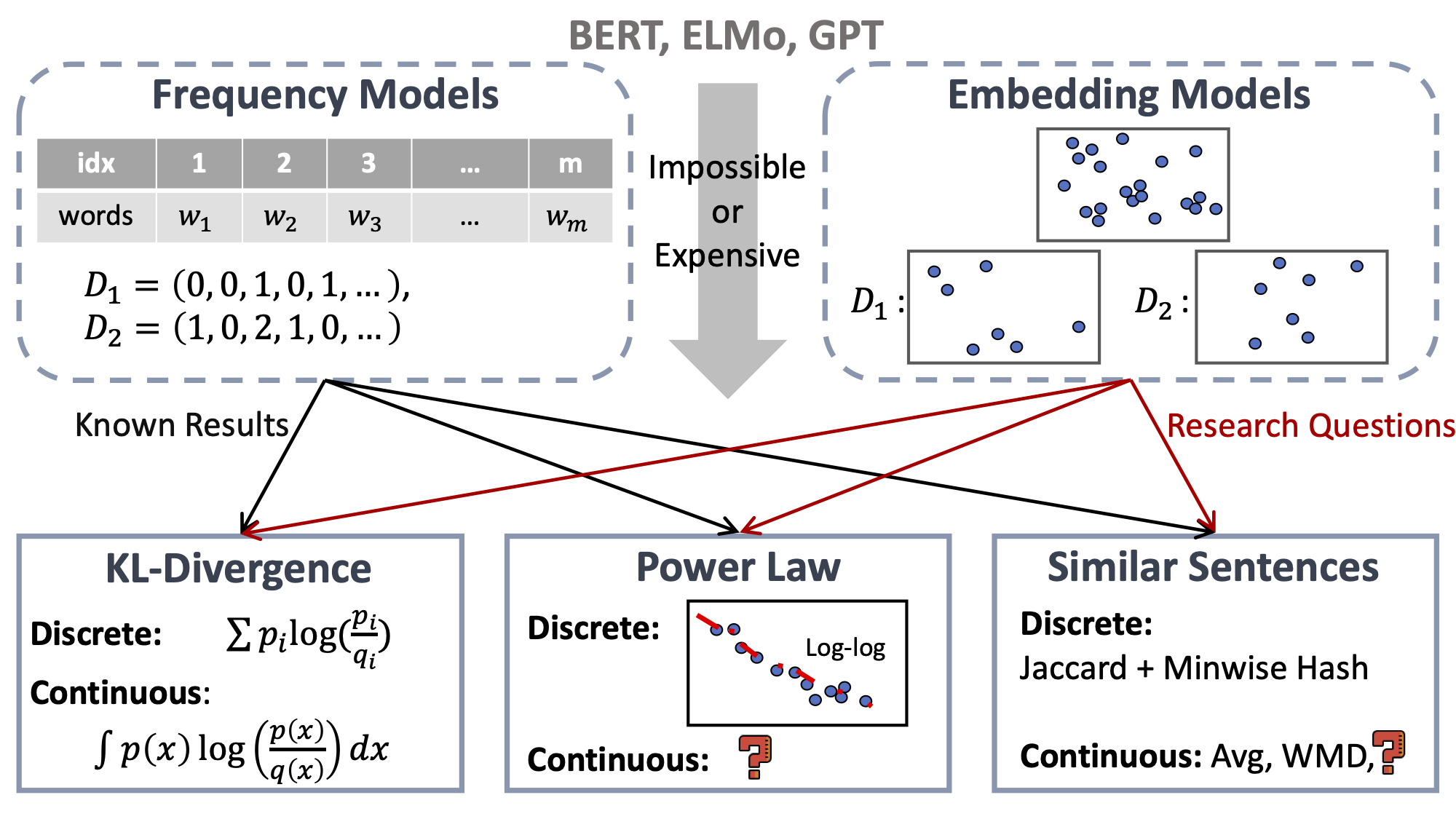}
\caption{{\small Remastering word-based toolkits for embedding models. Under word-frequency models, documents ($D_1$ and $D_2$) are represented by bag-of-word vectors, whereas under word embedding models, each document corresponds to a point cloud. We aim to answer three questions under embedding models: Q1. Can we design algorithms for estimating KL divergence? Q2. Does the distribution of the latent points exhibit any heavy tail (power law) behavior? Q3. Can we design efficient/effective algorithms to retrieve similar sentences? Note that it is difficult or expensive to perform these tasks at word level for powerful DL models such as BERT, ELMo, and GPT.}}
\label{fig:overview}
\vspace{-2mm}
\end{figure}

\section{Preliminaries and Related Work}


\subsection{Math background}
\label{sec:math-background}
\myparab{Notations.} For an arbitrary integer $k$, we let $[k] = \{1, \dots, k\}$. We let $\bD = \{D_1, D_2, \dots\}$ be a set of documents/sentences we are interested in modeling, and $W = \{w_1, \dots, w_m\}$ be the set of unique words that appear in the documents. We will sometimes use $D_j$ to denote the set of (unique) words that appear in document $j$. $f^{(j)}_i$ will denote the number of times that word $i$ appears in document $j$; when the document we refer to is clear, we shall write $f_i$ instead. Letting $n_j$ be the length (number of words) of document $j$, $p^{(j)}_i = f^{(j)}_i/n_i$ is the (empirical) occurrence probability of word $i$ in document $j$. 

\myparab{Entropy and divergence.} Let $\cald$ be a distribution on the discrete support $[k]$. Let $\tilde p_i$ be the probability mass for element $i$. The entropy of $\cald$ is $H(\cald) = - \sum_{i= 1}^k \tilde p_i \log \tilde p_i$. Let $\cald_1$ and $\cald_2$ be two different distributions. The KL divergence is defined as 
$\dkl(\cald_1 || \cald_2) = \sum_{i \in [k]} \tilde p_i \log\left(\frac{\tilde p_i}{\tilde q_i}\right),$
where $\tilde p_i$ ($\tilde q_i$) is the prob mass for $\cald_1$ ($\cald_2$). 

Let $D_j = \{D_{j, 1}, \dots, D_{j, n_j}\}$ ($j = 1, 2$) be a collection of i.i.d. samples from $\cald_j$. Recall we let $p^{(j)}_i$ be the empirical probability element $i$ appears in $D_j$. The empirical entropy of $D_j$ is $-\sum_{i = 1}^k p^{(j)}_i \log (p^{(j)}_i)$ and the empirical KL divergence of $D_1$ and $D_2$ is $\dkl(D_1 || D_2) = \sum_{i \in [k]}p^{(1)}_i \log \left(\frac{p^{(1)}_i}{p^{(2)}_i}\right)$.

\mypara{Continuous generalization.} The entropy and divergence for continuous variables can be defined accordingly. Let $X$ and $Y$ be two random variables in $\reals^{\ell}$ with probability density functions (pdfs) $f_X(\cdot)$ and $f_Y(\cdot)$ respectively. The differential entropy of $X$ is $H(X) = - \int f_X(x) \log f(x) dx$. The KL divergence between $X$ and $Y$ is 
\begin{align}\label{eqn:renyid}
    \dkl(X || Y) = \int f_X(x) \log \left( \frac{f_X(x)}{f_Y(x)}\right)dx.
\end{align}
The Rényi-$\alpha$ divergence is  
\begin{equation}
D_{R_{\alpha}}(X || Y) = \frac{1}{1-\alpha}\log\left(\int_{x }\frac{f_X(x)^\alpha}{f_Y(x)^{\alpha-1}}dx\right)
\label{eq:renyi-original}
\end{equation}
where $\alpha \in \mathbb{R} \setminus 1$ \cite{renyi_original}. 
One can see that $\lim_{\alpha\rightarrow 1} R_{\alpha}(X||Y) = \dkl(X||Y),$ which will be used in our estimation~\cite{kl_divergence_original}. 

We note that \emph{(i)} while differential entropy is a natural generalization of entropy, it sometimes can be negative and exhibit counter-intuitive behaviors \cite{cover_differential_entropy} so this paper focuses on only estimations of KL divergence. \emph{(ii)} Unlike discrete categorical distributions, there is no obvious generalization for the empirical entropy or KL-divergence. So our goal is to build an estimator that convergences the the ground-truth divergence as the sample size increases. 

\myparab{Zipf's law.} Zipf's law~\cite{zipf_1935, zipf_original} has a number of formulations for the occurrence probability $f(r)$ of the $r$-th most frequent item in a multinomial distribution. Most powerfully, it has been generalized as
\begin{equation}
f(r) = \frac{c}{(\beta + r)^\alpha},
\label{eq:zipf-mandelbrot}
\end{equation}
where $r$ is the rank, and $c$, $\alpha$, and $\beta$ are constants \cite{mandelbrot_1, mandelbrot_2, zipf_review}. This generalization is sometimes called the Zipf-Mandelbrot law and is equivalent to the generic form of Zipf's law when $\beta = 0$. Mandelbrot found the closest fit to natural language with $\alpha \approx 1$ and $\beta \approx 2.7$~\cite{zipf_review}. Note also that the Zipf distributions are heavy tailed, i.e., exhibiting a power law.

\subsection{Frequency-based models}
\label{sec:frequency-based-models}
Recall that in a word frequency model, each sentence or document is associated with a vector $v \in \reals^m$, where $m = |W|$ and $v_i$ is the number of occurrence of word $i$. These models usually assume that each word from the same document $D_j$ is sampled i.i.d. from the same underlying (multinomial) distribution $\cald_j$ on the support $W$.


\myparab{Power law distribution.} The frequencies of words in natural languages usually follow a power law distribution. Let $D$ be a document/sentence such that $p_i$ is the occurrence probability for word $i$. Let $i _1, i_2, \dots, i_m$ be a permutation on $[k]$ such that $p_{i_1} \geq p_{i_2} \geq \dots p_{i_m}$. Then $\{p_{i_j}\}_{j \leq m}$ can usually be modelled by a power law distribution (Eq.~\ref{eq:zipf-mandelbrot})~\cite{zipf_1935,zipf_original,zipf_review}. This observation remains valid when we consider word frequencies at a cruder granularity (e.g., a collection of books written in a given year, a collection of newspaper articles spanning multiple years \cite{zipf_brown_2}).

\mypara{Applications:} Zipf's law in language serves as a foundation for generative modeling of natural languages, statistical linguistic analysis, and efficient design of NLP algorithms~\cite{zipf_pareto} propose using Zipf's law to determine the parameters of a Pareto distribution. It can also be used to determine an appropriate vocabulary size where infrequent words need to be dropped to save computation \cite{zipf_applications}, as is often the case in DL with word-level recurrent language models \cite{LM_discarding_infrequent}. Finally, Zipf's law has been used to argue the effectiveness of language models at generating ``natural" text \cite{zipf_fitting_natural_language}. 

\mypara{Latent variable explanation:} One possible explanation of the pervasiveness of Zipfian and near-Zipfian is a latent variable in the data~\cite{zipf_latent}. The basic idea is that Zipf's Law holds when the frequencies for the observed variables (words in the case of text data) vary by several orders of magnitude, i.e., the distribution has a ``broad" range of frequency values. \cite{zipf_latent} argues that ``narrow'' distributions, meaning those with more balanced frequencies, are typical and that combining a series of narrow distributions with different means can result in the overall broad distribution observed in Zipf's law. 

In the case of text, this latent variable is taken to be part of speech. For many parts of speech, such as articles and pronouns, there are relatively few words that all occur with high frequency. For other parts of speech, like nouns and adjectives, there are many words and each is likely to occur with relatively low frequency. The frequency distribution across individual parts of speech does not follow Zipf's law (except for verbs, although these can be further segmented \cite{zipf_latent}), but when taken together the text as a whole exhibits the broad distribution that characterizes Zipf's law~\cite{zipf_latent}. 

\myparab{Estimation of language entropy and divergence.} Estimations of language entropy/divergence enjoy  widespread usage in a variety of applications, including reinforcement learning~\cite{kl_reinforcement}, feature selection~\cite{kl_feature_selection}, identification of discriminating terms in text~\cite{kl_discrim}, spam detection~\cite{kl_spam}, text classification \cite{kl_classification_symmetric}, and authorship identification~\cite{kl_div_authorship,kl_div_authorship_search,kl_authorship_1}.
This simple technique has been shown to outperform supervised methods such as Bayesian networks, support vector machines, and other standard  methods~\cite{kl_div_authorship,kl_div_authorship_search}. 

Under the word-frequency model, in which each word in a document is treated as an i.i.d. sample from an unknown distribution, estimating entropy/divergence is straightforward because we may merely use empirical estimates, which enjoy elegant properties derived from elementary statistical theory. For example, the estimator converges to the ground-truth when the sample sizes (document lengths) converge to infinity. 

\myparab{Similarity search.} A common information retrieval task is to efficiently compute the similarity between two documents, and find near neighbors of a query document. The most widely used document similarity measure is the Jaccard similarity, defined as 
$
J(D_{j_1}, D_{j_2}) = \frac{|D_{j_1} \cap D_{j_2}|}{|D_{j_1} \cup D_{j_2}|}.
$
Specialized data structures such as Minwise hash functions are designed to speed up the similarity computation, which has been widely used in industrial grade systems. 

Jaccard similarity has found its applications in a wide range of tasks such as information retrieval~\cite{jaccard_ir,jaccard_kmeans} and medicine~\cite{jaccard_segmentation,jaccard_permutation} because in part of its compute-efficiencies over other alternatives such as edit/Levenshtein distance~\cite{levenshtein}. 


\subsection{Word embedding models}
\label{sec:word-embedding-models}
In word embedding models, each word in $W$ is mapped into a high-dimensional space $\reals^{d}$ through a function $\psi(\cdot)$ so that two semantically related/similar words are mapped to two latent positions that are close to each other. For example, word2vec builds a probability model $\Pr[w' \mbox{ near } w \mid w]$, i.e., the probability that word $w'$ appears in a window centered at $w$. The model learns an embedding $\psi(w)$ together with a probability mass function $f$ such that $f(w', \psi(w))$ approximates $\Pr[w' \mbox{ near } w \mid w]$. We may view the function $f(\cdot, \psi(w))$ as the one indexed by $\psi(w)$ such that when two words are related, $\psi(w)$ and $\psi(w')$ are close, which also implies that $f(\cdot, \psi(w))$ and $f(\cdot, \psi(w'))$ are close. There are two methods of training associated with word2vec: continuous bag-of-words (CBOW), which predicts a word based on its context, and continuous skip-gram, which predicts context words based on a target word \cite{word2vec}. As pre-trained embeddings are easy to use and successfully captured several useful relationships among words, the word2vec embeddings were quickly adopted for a variety of tasks and inspired other pre-trained embeddings based on different algorithms or datasets, such as GloVe \cite{glove} and fastText \cite{fasttext}.

Recent work using DL has focused on contextual word embeddings that depend on neighboring words rather than using a single global representation for each word~\cite{cove,elmo,bert}. These contextual approaches require loading a pre-trained model and often involve learning a ``task-specific" weighting~\cite{elmo} or ``fine-tuning" embeddings by adding a classification layer~\cite{bert}, making them more somewhat more expensive to deploy than global embeddings. 

%% file: methods.tex
\section{Our NLP Methodology}

\subsection{Research Questions}
Instead of treating each word as a categorical variable, we use word-embedding techniques to represent the words in a high-dimensional latent space. So now we may interpret each word in a document/sentence as an i.i.d. sample from a continuous distribution. We highlight two edges of using word embedding. 

\mypara{Deep representation.} Word embedding captures relationships between words that cannot be captured by frequency models. Let us consider a toy example of a related sentence search. Consider the query sentence $Q$ = ``Queen Elizabeth II of England is one of the longest ruling monarchs in history." and the database consisting of two sentences: $s_1$ = ``The rock band Queen is famous for songs like 'Bohemian Rhapsody'." and $s_2$ = ``King Louis XIV, former ruler of France, reigned more days than any other sovereign." Clearly, $Q$ is closely related to $s_2$ and largely unrelated to $s_1$, since $s_1$ is about a band and $s_2$ is also about long-reigning monarchs. The query sentence has 13 unique words, $s_1$ has 11, and $s_2$ has 14. We compute the Jaccard similarity $J(Q, s_1) = \frac{2}{22} \approx 0.091$, where both sentences share ``Queen" and ``is". We also compute $J(Q, s_2) = \frac{1}{26} \approx 0.038$, where both sentences share ``of". If we remove common stopwords, $Q$ and $s_2$ have a Jaccard similarity of zero. Clearly, this approach is too simplistic for the similar sentence problem as it relies on exact matching. We look to use word embeddings to allow for ``fuzzy" matches between two distinct words with similar semantic features.

\mypara{Transfer learning vs. generalization errors.}
Using existing word embedding models provides a balanced trade-off between their expressive power and their generalization error. Models such as fastText and GloVe are trained using huge datasets that aim to capture the co-occurrence of words. They do not account for long-term dependencies between words (which usually are application specific) so while these models are less powerful than recurrent models (BERT, ELMo etc.) that use context-specific representations, they are more robust against different application scenarios. In other words, knowledge ``pre-trained'' under the word embedding model can be easily transferred to other application scenarios. 

\myparab{Our research questions.}
Our major goal is to generalize algorithmic techniques and statistical analysis developed for word-frequencies models to word-embedding models. We specifically investigate three research questions. 


\mypara{Q1. Can we develop a statistically sound algorithm to estimate the divergence between two documents?} Let $D_1$ and $D_2$ be two documents sampled from two distributions $\cald_1$ and $\cald_2$ over $\reals^{\ell}$. Our goal is to build an unbiased estimator for the KL divergence between $D_1$ and $D_2$ by 
using independent samples from $\cald_1$ and $\cald_2$. 

We integrate two techniques developed recently, one algorithmic and the other computational, to design the estimator. First, we cast this problem as a non-parametric estimation problem (i.e., we do not make any distributional assumptions on either $D_1$ or $D_2$) and employ a nearest-neighbor (NN) based estimator developed by~\cite{renyi_estimation}  to find $\dkl(D_1, D_2)$. Next, standard $k$-NN libraries cannot properly handle the data scale we face so we apply a recently released approximate $k$-NN library \cite{faiss} to speed up our computation. 

\mypara{Q2. Can we confirm that the continuous distribution for the latent positions exhibits any heavy tail?} Let $f_{\cald}(\cdot)$ be the pdf (a continuous function) of the underlying distribution for generating the observed document. Our goal is to understand whether $f_{\cald}(\cdot)$ exhibits any power law structure. Our main discovery is that if we examine the points at the sentence level, where each sentence is represented by the mean of all the embedded words, and cluster the sentences, the size (i.e., the number of points) of each cluster follows a power law distribution. This implies that we may partition the latent spaces into multiple blocks so that each block contains related sentences. We also observe that in the latent space words are frequently close to other words with the same part of speech. This suggests that this characteristic, which has been suggested as an explanation for Zipf's law, may be observable from representations in the latent space \cite{zipf_latent}.

Our observation serves as a continuous counterpart for frequency models, where similar phenomena were found.

\mypara{Q3. Can we develop an efficient algorithm to search for similar documents?} We examine how we may generalize the Jaccard similarity for continuous variables. When we need to compute the similarities between two sets of continuous variables, this intuitively means that we need to define the ``distance'' between two clouds of points. 

There have been two standard approaches to solve the problem. One represents a cloud by the average of its constituent points, thereby reducing the similarity comparison between clouds to computing the distance between two average points. This method is computationally efficient but discards information by ``summarizing" each cloud and does not always yield intuitive results. A second method is an extension of the earth movers distance~\cite{emd} known as the word movers distance (WMD) \cite{wmd}. This method delivers more intuitive and interpretable search results at the cost of using significantly more resources. Here, we design a combinatorial algorithm that translates the problem to a set cover problem. Our approach balances well between the compute performance (e.g., it is embarrassingly parallel) and the scrutability (by retaining all of the information in the cloud, it can capture subtler relationships while remaining intuitive to users). Additionally, when run in succession on the same query, our algorithm identifies a better variety of similar sentences than others.

\subsection{Divergence Estimation}
Recall that we are interested in estimating the KL divergence between two distributions $\cald_1$ and $\cald_2$. We observe two documents $D_1$ and $D_2$ such that words in $D_i$ are i.i.d. samples from $\cald_i$ ($i = 1, 2$). Let  
$X_{1:N} = (X_1, \dots, X_N)$ be words (samples) from $D_1$ and $Y_{1:M} = (Y_1, \dots, Y_M)$ be words from $D_2$. Our goal is to find the KL divergence between $\cald_1$ and $\cald_2$ based on $X_{1:N}$ and $Y_{1:M}$. 

Let $f_1(\cdot)$ and $f_2(\cdot)$ be the pdfs of the distributions $\cald_1$ and $\cald_2$ respectively. If we knew the exact form of $f_i(\cdot)$, computing the divergence would be straightforward. Because we only observe independent samples from $\cald_1$ and $\cald_2$, we must rely on these to approximate the pdfs. One standard approach is to use a kernel estimator~\cite{wasserman2006all} but this method will introduce long range dependencies between points (e.g., estimating $f_i(x)$ depends on points that are far from $x$) so it is difficult to parallelize. 

Instead of using kernel estimators, we rely on the estimator proposed by P{\'o}czos et al.~\cite{renyi_estimation}. This new estimator relies on only the nearest neighbors of a specific point $x = \psi(w)$ to determine $f_i(x)$. Intuitively, let $x' = \psi(w')$ be the $k$-th nearest neighbor of $x$. We then may use $\|x - x'\|$ to determine the density around $x$: when $\|x - x'\|$ is small, it means the ball centered at $x$ is densely packed and thus the density function at $x$ should be large. Similarly, when $\|x - x'\|$ is large, the density value should be small. 

P{\'o}czos et al.~\cite{renyi_estimation} showed that this algorithm is effective in estimating the Rényi-$\alpha$ divergence. Since the Rényi-$\alpha$ divergence converges to KL divergence with $\alpha \rightarrow 1$, we may use P{\'o}czos et al.'s estimator to find the KL divergence. 

Specifically, we estimate the sum inside the logarithm in Eq.~\ref{eqn:renyid} (i.e., $\delta \triangleq \int_x \frac{f^{\alpha}_1(x)}{f^{\alpha-1}_2(x)}dx$) by 
\begin{equation}
\hat{\sigma}_k =\frac{1}{N} \sum_{i=1}^N\left(\frac{(N-1)\rho_k(X_i)}{M\upsilon_k(X_i)} \right)^{1-\alpha}B_{k, \alpha}
\label{eq:sigma_estimation}
\end{equation}
where $k$ is a hyperparameter, indicating which neighbor should be used for the estimation, $\rho_k(x)$ is the Euclidean distance from $x$ to its $k$-th nearest neighbor in $X_{1:N} \setminus x$,  $\upsilon_k(x)$ is the Euclidean distance from $x$ to its $k$-th nearest neighbor in $Y_{1:M} \setminus x$, and $B_{k, \alpha} = \frac{\Gamma (k)^2}{\Gamma (k - \alpha + 1)\Gamma (k + \alpha - 1)}$ for $k > |\alpha - 1|$~\cite{renyi_estimation}.

Then we have $\hat{D}_{R_{\alpha}, k}(\mathcal{D}_1, \mathcal{D}_2) = \frac{1}{\alpha - 1}\log (\hat{\sigma}_k)$. Note that both the divergence estimate and $\sigma$ estimates depend on the choice of $k$~\cite{renyi_estimation}. 
Finally, we set $\alpha$ be a value close to 1 to approximate the KL-divergence, i.e., let $\epsilon  = 10^{-5}$ and estimate
\begin{equation}
\hat{D}_{kl, k}(\mathcal{D}_1, \mathcal{D}_2) = \frac{1}{2}(\hat{D}_{R_{1 + \epsilon}, k}(\mathcal{D}_1, \mathcal{D}_2) + \hat{D}_{R_{1 - \epsilon}, k}(\mathcal{D}_1, \mathcal{D}_2)).
\label{eq:kl_estimation}
\end{equation}
This $\epsilon$ was chosen to be small but not small enough to cause a divide by zero error. We approach from both above and below in an attempt to best approximate the value at the limit.

\subsection{Zipf's law}
To analyze Zipf's law in the latent space, we wanted to see if the latent variable of part of speech was captured in the embeddings. Assuming this explanation has some validity, identifying parts of speech in the latent space indicates that the underlying mechanism of Zipf's law is present and suggests that we do not simply discover a new power law distribution. In other words, if we show that the conditions that cause a heavy-tailed distribution in the categorical space are also present in the latent space, this is evidence that an analogous heavy-tailed distribution should exist in the latent space. 

To test this, we will first assign each token a part of speech using a standard library. As each embedding is assigned a single part of speech even though a single English word may have multiple parts of speech, some results will constitute lower bounds because they could be improved by a different way of tagging each word. We will examine the part of speech composition of neighborhoods by both counting the number of neighbors with the same part of speech for each point and by visualizing the results using dimensionality reduction techniques. 

Observing that many sentences in a body of work or a corpus are likely to share similar themes and structures, we attempt to identify a heavy tail distribution by clustering sentences using k-means. Specifically, we take the average embedding of words in each sentence and perform k-means clustering on these averages. After clustering, we compare the number of sentences in each cluster with the rank of that cluster in terms of number of sentences. We expect that there will be a few clusters with very many sentences and many clusters with few sentences. 



\subsection{Similarity Search}
\myparab{Set-cover based search algorithms.} Let $\mathcal{T} = \{s_1, s_2, \dots, s_d\}$ be a database of $d$ sentences, 
where each sentence $s_i$ is a set of words: $s_i = \{s_{i, 1}, s_{i, 2}, \dots, s_{i, n_i}\}$ so that $n_i$ is the length of sentence $i$ and each $s_{i, j} \in W$. Let $W_{\mathcal{T}}$ be the set of words in $\mathcal{T}$, i.e., $W_{\mathcal{T}} = \{s_{i, j}\ |\forall i \in \{1, \dots, d\}, j \in \{1, \dots, n_i\} \}$ and $W_{\mathcal{T}} \subseteq W$. Recall that $\psi(w)$ is the embedding for the word $w \in W$. For $w \in W$, we can define $\mathbf{Ball}_{\mathcal{T}, r}(w)$ as the set that contains exactly $r$ nearest neighbors of $w$. Let $\mathbf{Ball}_{\mathcal{T}, r}(w) = \{w_{i_1}, \dots, w_{i_{r}}\}$ be the $w_i \in W_{\mathcal{T}} /\{w\}$ such that 
$\|\psi(w) - \psi(w_{i_{j_1}})\| \leq \|\psi(w) - \psi(w_{i_{j_2}})\|$ if and only if $j_1 \leq j_2$. Similarly, for a set of words $A = \{w_{a, 1}, \dots, w_{a, n_a} \}$ we can define $\mathbf{Ball}_{\mathcal{T}, r}(A) = \cup_{w_{a, i} \in A} \mathbf{Ball}_{\mathcal{T}, r}(w_{a, i})$. In words, the ball of a set is the union of the balls for each item in the set.

Let $Q = \{w_{q_1}, w_{q_2}, \dots, w_{q_{n_q}}\}$ be a query sentence of length $n_q$ such that $w_{q_j} \in W$. The similarity search problem is to find the $k$ sentences in $\mathcal{T}$ that are the ``most similar'' to the query sentence $Q$. We find a set of sentences $S \subseteq \mathcal{T}$ of size $t$ such that $\left(\bigcup_{s \in S}\{w \in s\}\right)$ $\cap \mathbf{Ball}_{\mathcal{T}, r}(Q)$ is maximized (i.e., choose $t$ sentences that best ``cover" the words in $Q$). Recognizing that words need not be identical to convey the same or similar meaning, we leverage word embeddings to identify semantically similar words to those in $Q$ and include them as target words to be covered. By allowing these ``fuzzy" matches that are impossible with categorical representations alone, we increase the likelihood to identify related sentences.

\begin{wrapfigure}{R}{0.25\textwidth}
  \vspace{-20pt}
  \begin{minipage}{0.25\textwidth}
    \begin{algorithm}[H]
      \caption{\small A simple algorithm for finding similar sentences. $Q$ is the query sentence, $\mathcal{T}$ is the database of sentences, $E$ is the set of English stopwords, $t$ is the number of sentences to return, $r$ is the number of nearest neighbors to consider, and $\rho$ is the length penalty.}
      \label{alg:similar_sentence}
      \begin{algorithmic}
          \REQUIRE $Q, \mathcal{T}, E, \rho, t \geq 1, r \geq 0$
          \STATE $S \leftarrow \emptyset$
          \STATE $U \leftarrow \mathbf{Ball}_{\mathcal{T}, r}(Q)$
          \STATE $U \leftarrow U \setminus E$
          \FOR{$j$ from $1$ to $t$}
          \STATE $s_{i^*} \leftarrow \argmax_{s_i \in \mathcal{T}}\left(\frac{|U \cap s_i|}{|s_i|^p}\right)$
          \STATE $U \leftarrow U\setminus (U \cap s_{i^*})$
          \STATE $\mathcal{T} \leftarrow \mathcal{T} \setminus s_{i^*}$
          \STATE $S \leftarrow S \cup s_{i^*}$
          \ENDFOR
          \RETURN $S$
      \end{algorithmic}
    \end{algorithm}
  \end{minipage}
  \vspace{-10pt}
\end{wrapfigure}

Specifically, consider the set of elements to cover $U$ as $\mathbf{Ball}_{\mathcal{T}, r}(Q)$, that is, the union of the $r$ nearest neighbors of each word in the query sentence. Each sentence $s_i \in \mathcal{T}$ can be viewed as a (possibly empty) set of elements (words) in $U$ and is assigned a cost $c(s_i)$ of one. Our set-cover based search problem aims to find a solution set $S \subseteq \mathcal{T}$ such that $\sum_{s_i \in S}c(s_i) = t$ and $|\left(\bigcup_{s_i \in S} \{w_i \in s_i\}\right) \cap U|$ is maximized. Although the set-cover based search problem is equivalent to set cover problem and thus is NP-complete, there exist greedy algorithms with a constant approximation ratio \cite{set_cover_book}.

\myparab{Advantages of Set Cover.}
\emph{(i)} The algorithm explicitly selects varied sentences by removing words ``covered" by previously suggested sentences from being rewarded in subsequent recommendations. This recommendation diversity is especially important for sentence recommendation in certain applications such as creative writing, where the intent is to inspire the user to continue writing. Repeatedly recommending sentences that vary by only a few words is unlikely to sufficiently change context in a way that produces new ideas for a writer, which is a drawback that categorical approaches such as Jaccard similarity suffer from.  \emph{(ii) } Our algorithm balances performance and scrutability, which cannot be achieved by other solutions such as earth-moving distance based ones~\cite{speeding_wmd} or simple averaging. By leveraging high performance indices on GPUs such as \texttt{faiss} ~\cite{faiss}, nearest neighbor look ups may be done extremely efficiently. For each query, nearest neighbors only need to be looked up a single time and the rest of the operations for identifying the most similar sentence rely on efficient set operations, resulting in a highly scalable solution. The set cover approach is scrutable because one can easily construct an explanation for why each sentence was recommended, for instance by mapping words from the suggested sentence to the query sentence. In situations where this is especially important, it could be aided by showing the nearest neighbors of each word in the query sentence. This makes parameter tuning and debugging easier in addition to providing users with an intuitive understanding of the results. 

\myparab{Improved Heuristics} 
To improve upon the na\"iive greedy algorithm for the set cover problem, we consider a few heuristics. First, we recognize that neither long sentences nor very short sentences are likely to be useful. Some long sentences may contain many of the words in $\mathbf{Ball}_{\mathcal{T}, r}(Q)$, but if we do not penalize them we are likely to see sentences with a low signal to noise ratio. To address this, we select the sentence $s_i$ that maximizes $\frac{|U \cap s_i|}{|s_i|^{\rho}}$ where $\rho$ is a tunable parameter set to 0.5 for our experiments. Similarly, we remove all sentences with fewer than five tokens from consideration as they are too short to provide useful information. This can be accomplished either by removing short sentences from the database or setting the score to zero. Second, to remove more noise, we delete common English stopwords $E$ from $U$ using a standard list from \texttt{nltk} \cite{nltk}, as these are unlikely to help us finding similar sentences.

%% file: experiments.tex
\section{Experiments and Results}

\begin{table}[t]
\renewcommand{\arraystretch}{1}
\begin{center}
\small
\caption{{\small Number of books labeled correctly for various KL divergence tasks, where baseline is the original relative frequency based calculation. As the document size grows, the performance of our estimator improves over the categorical approach.}}
\begin{tabular}{lc||ccccccc}
\hline
\hline
&  & & \multicolumn{6}{c}{\textit{Value of k}}\\
Task & Size & Baseline & 3 & 5 & 10 & 25 & 50 & 100\\
\hline
Authorship & 3k & \textbf{28} & 23 & 17 & 13 & 11 & 8 & 7 \\
Reading level & 3k &  \textbf{36} & 32 & 32 & 34  & 24 & 22 & 21\\
Genre & 3k &  68 & \textbf{69} & 68 & 62 & 35 & 17 & 11\\
Authorship & 10k &  \textbf{21} & 16 & 13 & 11 & 7 & 3 & 3 \\
Reading level & 10k &  18 & \textbf{21} & 18 & 20 & 19 & 19 & 19\\
Genre & 10k &  \textbf{45} & 41 & 43 & 44 & 34 & 18 & 10\\
Authorship & 20k &  \textbf{16} & 15 & 10 & 6 & 3 & 2 & 2 \\
Reading level & 20k &  22 & \textbf{24} & 23 & \textbf{24} & 18 & 18 & 18\\
Genre & 20k &  \textbf{42} & \textbf{42} & \textbf{42} & \textbf{42} & 41 & 37 & 25\\
Authorship & 32k &  18 & \textbf{20} & 16 & 11 & 8 & 7 & 6 \\
Reading level & 32k &  18 & \textbf{25} & 17 & 19 & 20 & 16 & 16\\
Genre & 32k &  \textbf{39} & 38 & \textbf{39} & \textbf{39} & 38 & 35 & 14\\
\end{tabular}
\label{tab:kl-results}
\vspace{-2mm}
\end{center}
\end{table}

\input{data.tex}

\subsection{Divergence Estimation}
\label{entropy_estimation_experiment}
To test our divergence estimate, we focus on the Gutenberg dataset. We can divide these books across three dimensions: authorship, genre, and reading level. Across the 79 works, we have 18 authors with more than one entry, giving us 19 categories: one for each author plus a ``unique" category with 34 entries. For genre, we divide the works into ``fiction," ``non-fiction", ``poetry", and ``plays" so each would have a varied and substantial group of books. Lastly, we create groups corresponding to a pre-high school reading level, a high school reading level, and a college reading level. Generally, pre-high school books tend to be shorter stories and college books are particularly complex or long or are relatively dense non-fiction. 

\begin{table}[t]
\renewcommand{\arraystretch}{1}
\small
\begin{center}
\caption{{\small Average percentage of neighbors with the same part of speech as the target word. Even when considering 50 neighbors, our larger datasets find \char`\~ 70\% are of the same part of speech, another indicator that part of speech data is captured in word embeddings.}}
\begin{tabular}{c||ccccc}
\hline
\hline
& \multicolumn{5}{c}{Number of Neighbors ($k$)}\\
Dataset & 3 & 5 & 10 & 20 & 50\\ 
\hline
Gastby & 68.53 & 65.81 & 61.71 & 56.76 & 50.60\\
Reuters & 83.07 & 82.66 & 81.82 & 80.60 & 78.65\\
Brown & 75.72 & 74.83 & 73.32 & 71.29 & 67.96\\
Gutenberg & 77.45 & 76.67 & 75.57 & 74.15 & 71.87\\
RACE & 79.10 & 78.09 & 76.72 & 75.12 & 72.78\\
\end{tabular}
\label{pos_nbrs}
\vspace{-2mm}
\end{center}
\end{table}


To test our estimate, we first compute $\dkl(P || Q)$ for each pair of books. Then, for each category $C$, we calculate $\dkl(P || C)$ by averaging $\dkl(P || c)$ for each $c \in C$. We can then label $P$ with the category that minimizes this average. We note that when $N$ and $M$ (in Eq.~\ref{eq:sigma_estimation}) differ by orders of magnitude, the ratio between them dominates the $\rho$ and $\upsilon$ terms, and could potentially result in negative values in our KL divergence estimation. This is not an issue of lack of data as it depends on the ratio between $N$ and $M$. To avoid this problem, instead of using the full text we sample the same number of tokens at random from each document, discarding shorter documents. We examine four sample sizes: 3000 words from 76 books, 10,000 words from 51 books, 20,000 words from 46 books, and 32,000 words from 42 books. Samples of size $n$ were generated by tokenizing each book, removing stopwords, shuffling, and returning the first $n$ tokens. 

\begin{figure*}[t]
\centering
\includegraphics[width=\textwidth]{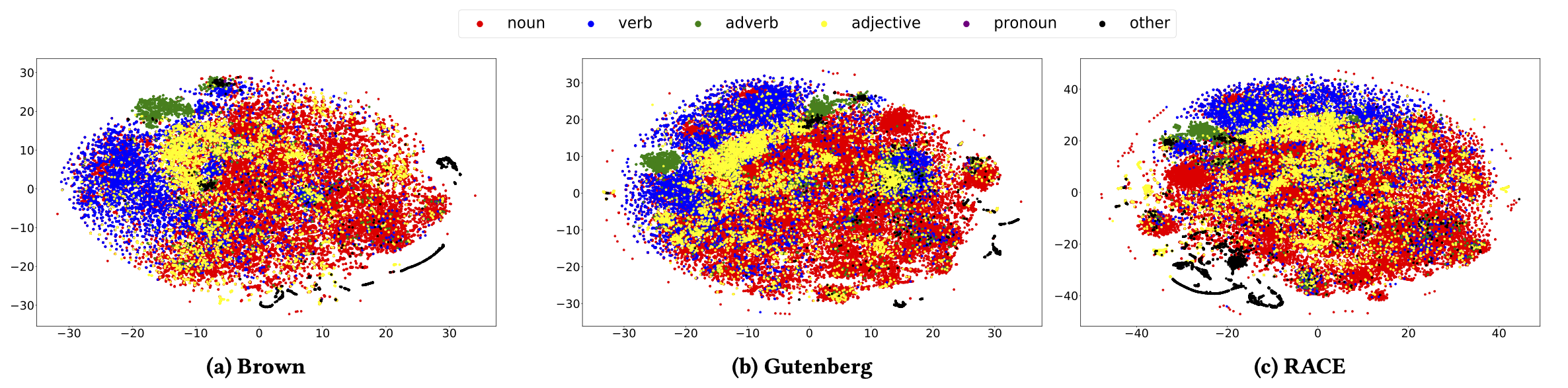}
\caption{\small{t-SNE plots of unique words, colored by part of speech. We show only the three largest datasets here; results for Gatsby and Reuters can be found in Appendix \ref{app:part-of-speech}. Even in a reduction from 300 dimensions to two, it appears that words are frequently close based on part of speech. This suggests the latent variable which may account for Zipf's law in the categorical case is also observable in the word embeddings.}}
\label{fig:tsne}
\end{figure*}

The results with different text sizes across different values of $k$ for entropy estimation can be found in Table \ref{tab:kl-results}. We observe that the ``unique" category for authorship is almost never chosen, which accounts for for the relatively low accuracy on the authorship attribution task. For smaller documents, the estimator achieves similar performance on the reading level and genre tasks for small values of $k$. As the sample size increases, the estimator outperforms the baseline on each task for different values of $k$. It appears that $k=3$ achieves the best results among the estimators for most tasks, although $k=10$ often sees comparable or better results for the genre and reading level tasks. We observe that occasionally higher values of $k$ correctly make classification that the lower, generally more accurate values of $k$ do not. 
These results suggest that while the results are usually robust against different $k$'s, sometimes fine-tuning task-specific $k$ may improve our algorithm's performance. 

\subsection{Zipf's Law}
\subsubsection{Part of Speech Analysis}
\input{part_of_speech.tex}

\subsubsection{Cluster Size}
For each dataset, we first get the set of sentences from either the dataset directly or using the $\texttt{nltk}$ sentence tokenizer. We then tokenize each sentence using $\texttt{nltk}$ and look up each token in the fastText embedding dictionary. Each sentence is represented by an average embedding of its component words; this average embedding is used as the feature set for a k-means classifier. To choose $k$, we inspected the five sentences closest to the mean and the five furthest from the mean to see if they shared a common thread. For instance, for the Gatsby dataset, one cluster was all short sentences containing the name "Gatsby," while another consisted entirely of short questions. We aim to find a suitably large $k$ such that the same theme was not split across multiple clusters. Small tweaks to $k$ and re-running the clustering algorithm again with a different random seed did not appear to have a major effect on our observation; for the robustness of our results with respect to the choice of $k$, see Appendix \ref{app:zipf-changes-k}. The choices of $k$ are as follows: 35 for Gatsby, 85 for Reuters, 75 for Brown, 100 for Gutenberg, and 95 for RACE.

\begin{figure*}[t]
\centering
\subfloat[Gatsby]{\includegraphics[width=1.25in]{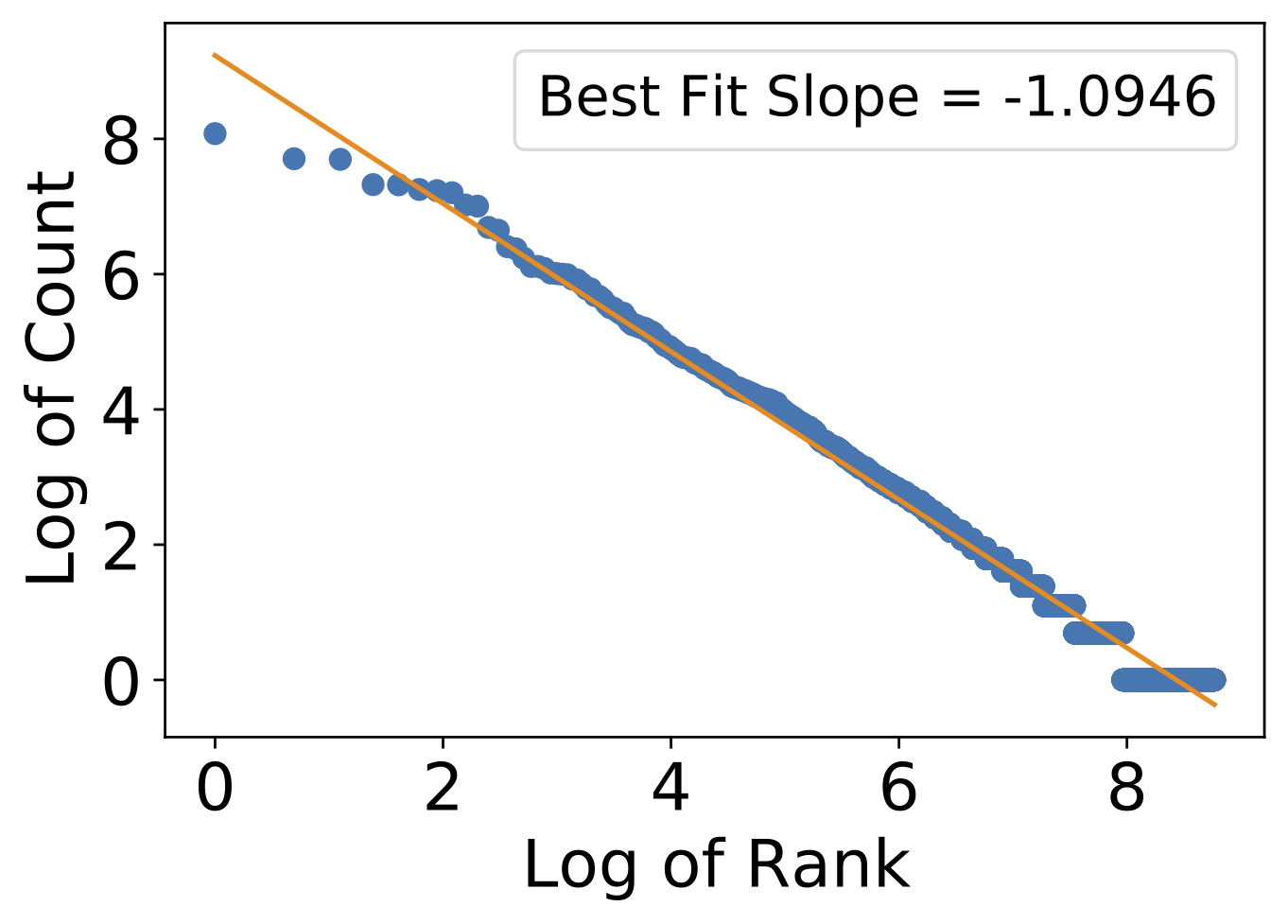}
\label{zipf_gatsby}}
\subfloat[Reuters]{\includegraphics[width=1.25in]{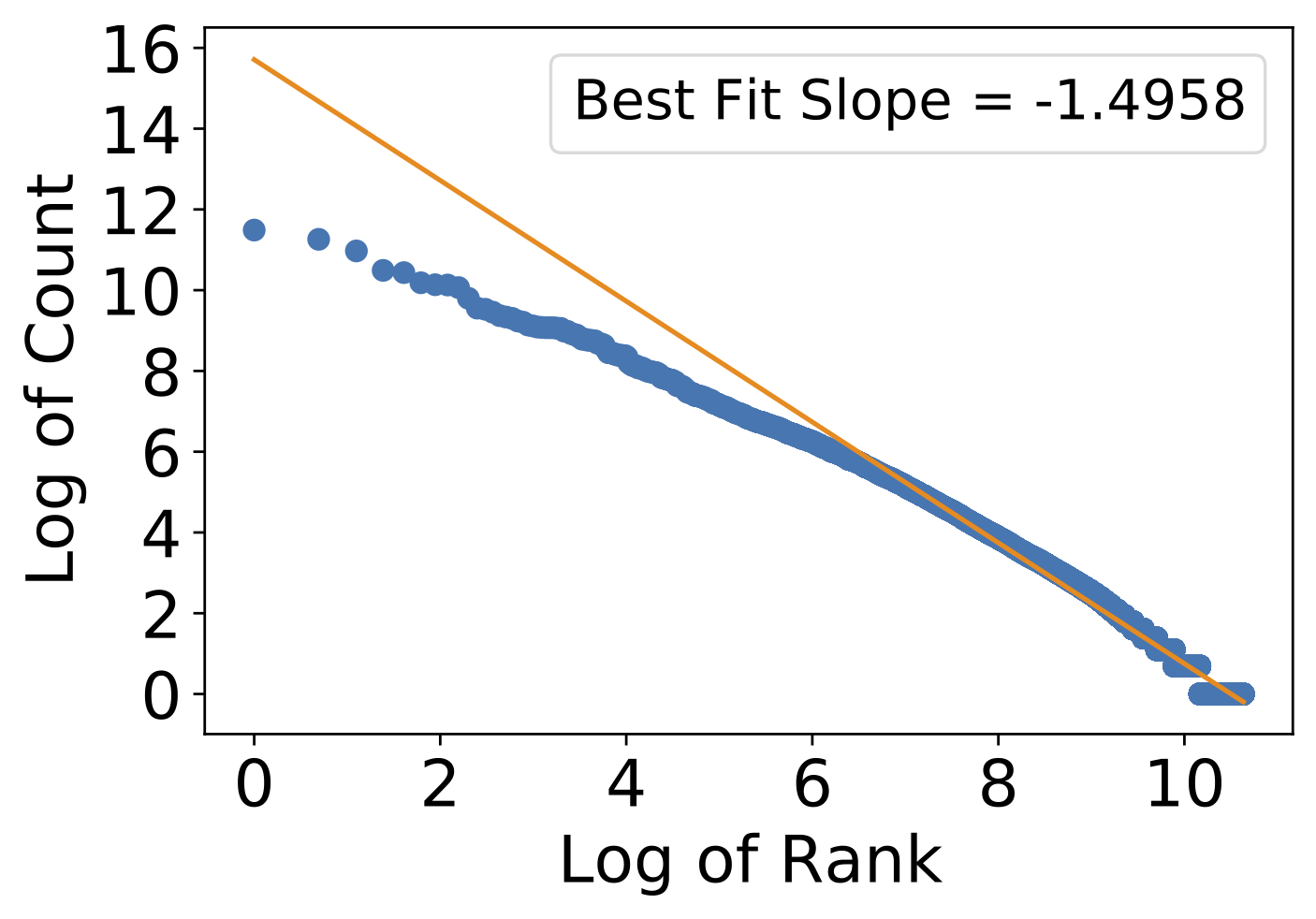}
\label{zipf_reuters}}
\subfloat[Brown]{\includegraphics[width=1.25in]{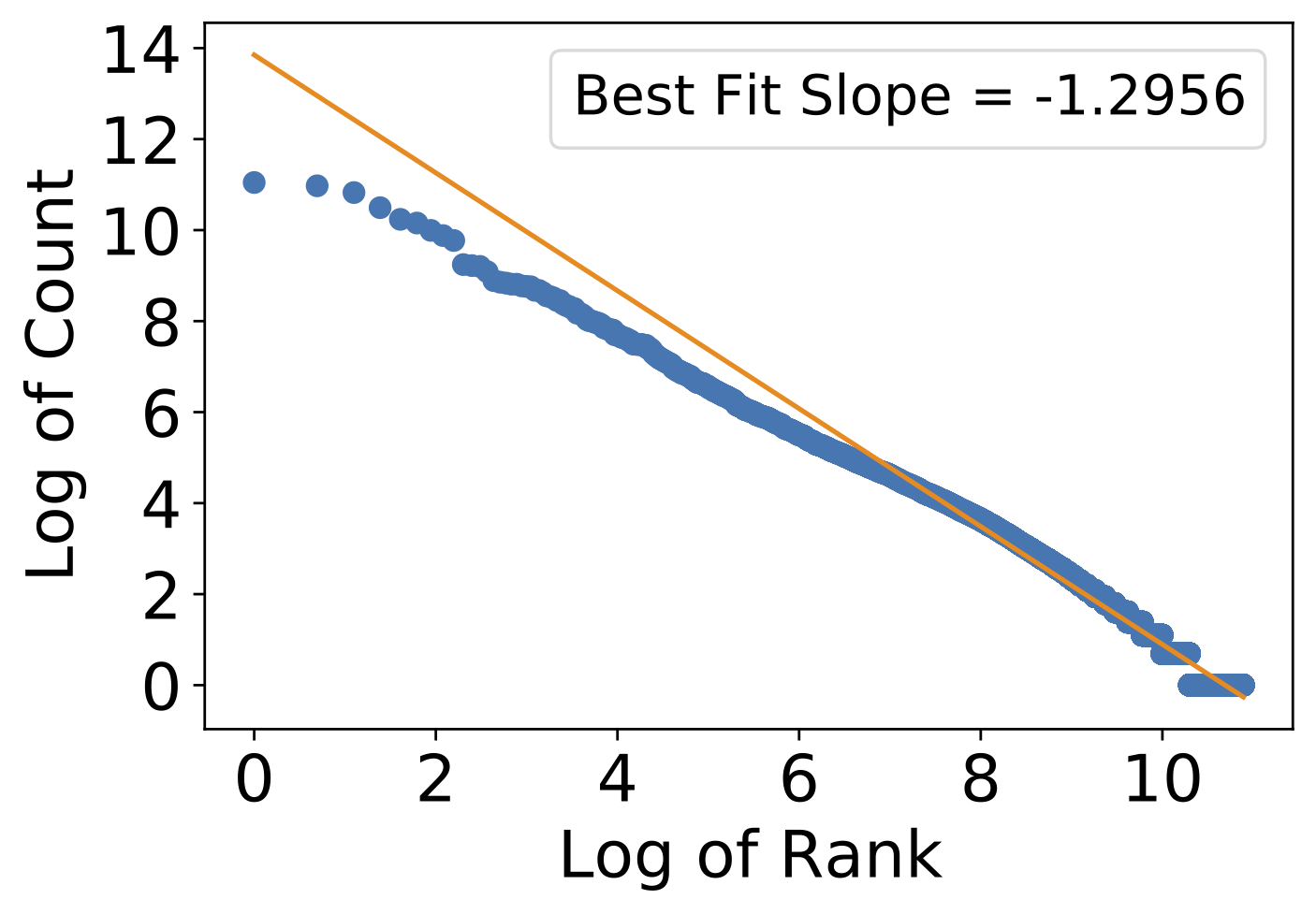}
\label{zipf_brown}}
\subfloat[Gutenberg]{\includegraphics[width=1.25in]{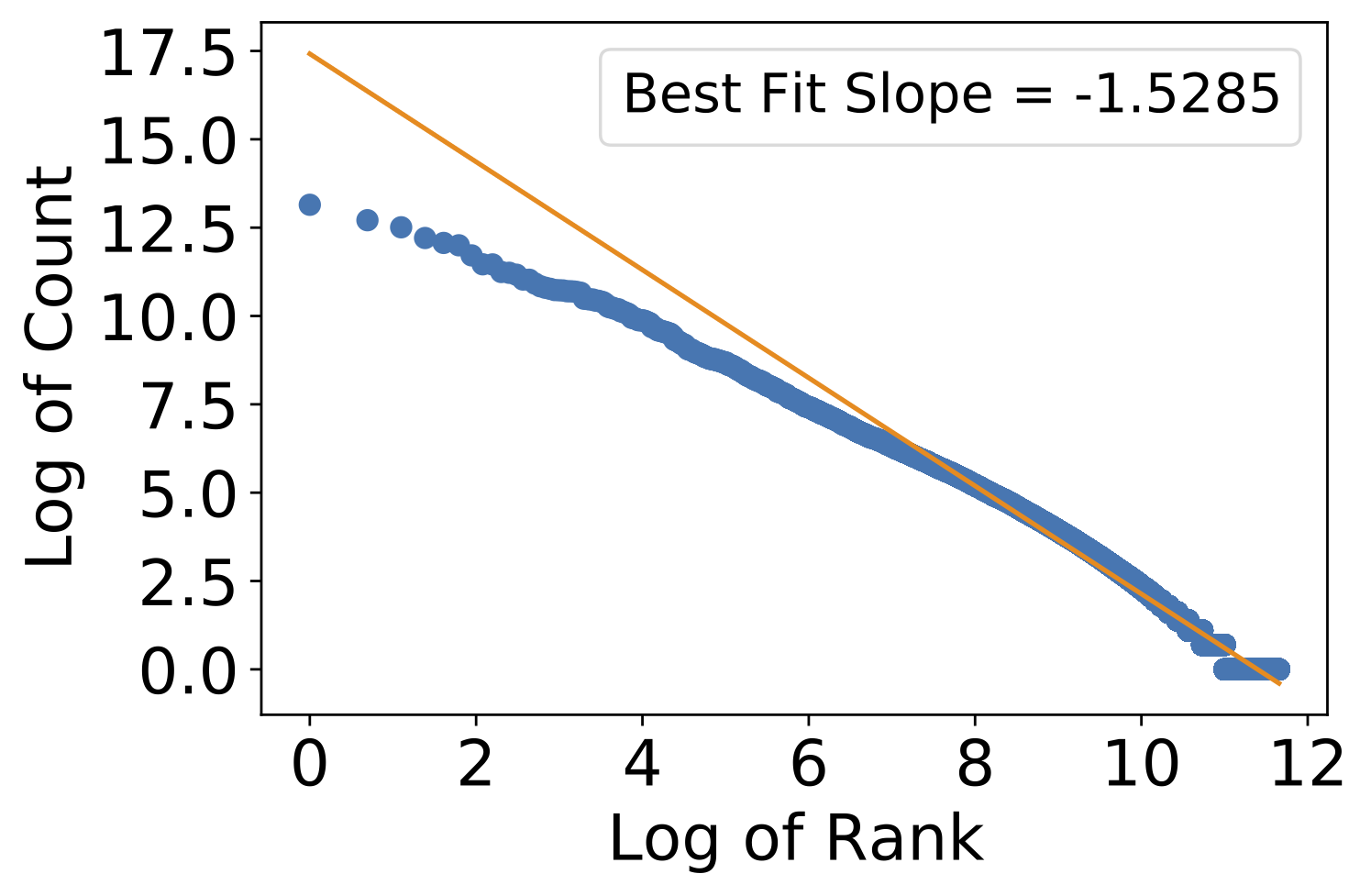}
\label{zipf_gutenberg}}
\subfloat[RACE]{\includegraphics[width=1.25in]{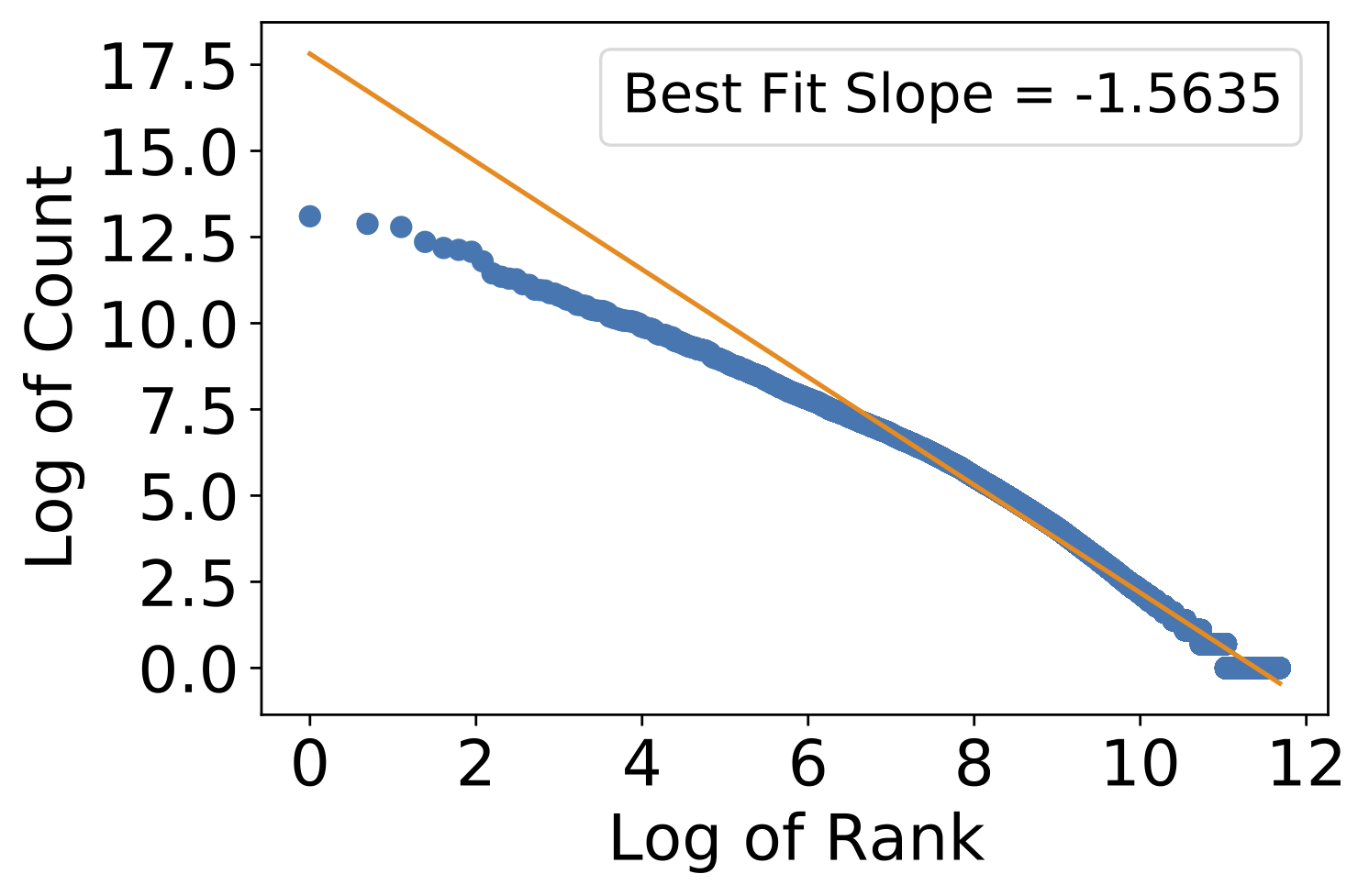}
\label{zipf_race}}
\hfil
\vspace{-0.1in}
\caption{\small{log-log plots of word count vs rank, with line of best fit. We can see that the Gatsby and Brown datasets fit a heavy-tailed distribution quite well and that Reuters, Gutenberg, and RACE fit less well but are still close to the slope of $-1$ expected for Zipf's law.}}
\label{fig:original_zipfs}
\end{figure*}

\begin{figure*}[t]
\vspace{-0.2in}
\centering
\subfloat[Gatsby]{\includegraphics[width=1.25in]{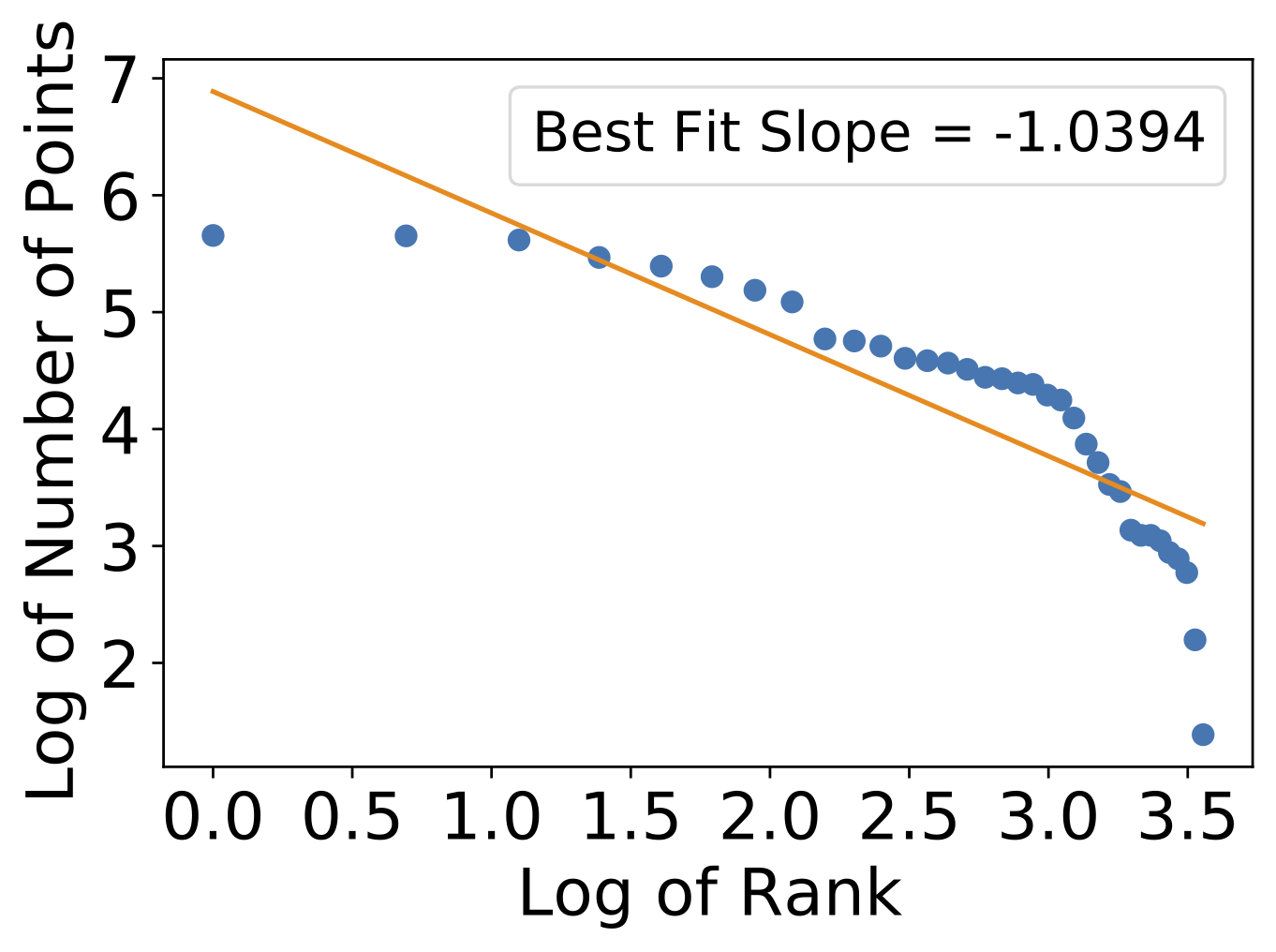}
\label{count_rank_gatsby}}
\subfloat[Reuters]{\includegraphics[width=1.25in]{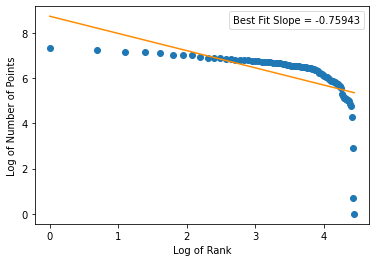}
\label{count_rank_reuters}}
\subfloat[Brown]{\includegraphics[width=1.25in]{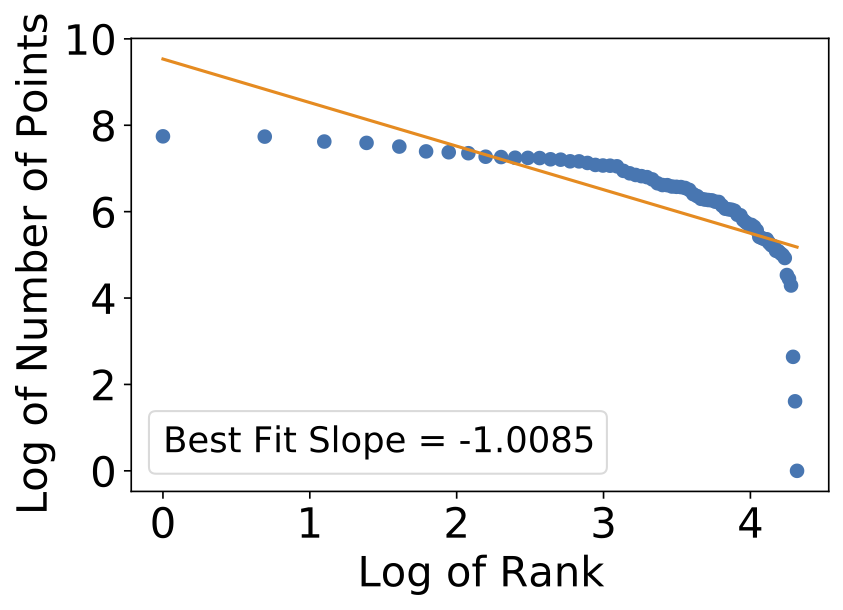}
\label{count_rank_brown}}
\subfloat[Gutenberg]{\includegraphics[width=1.25in]{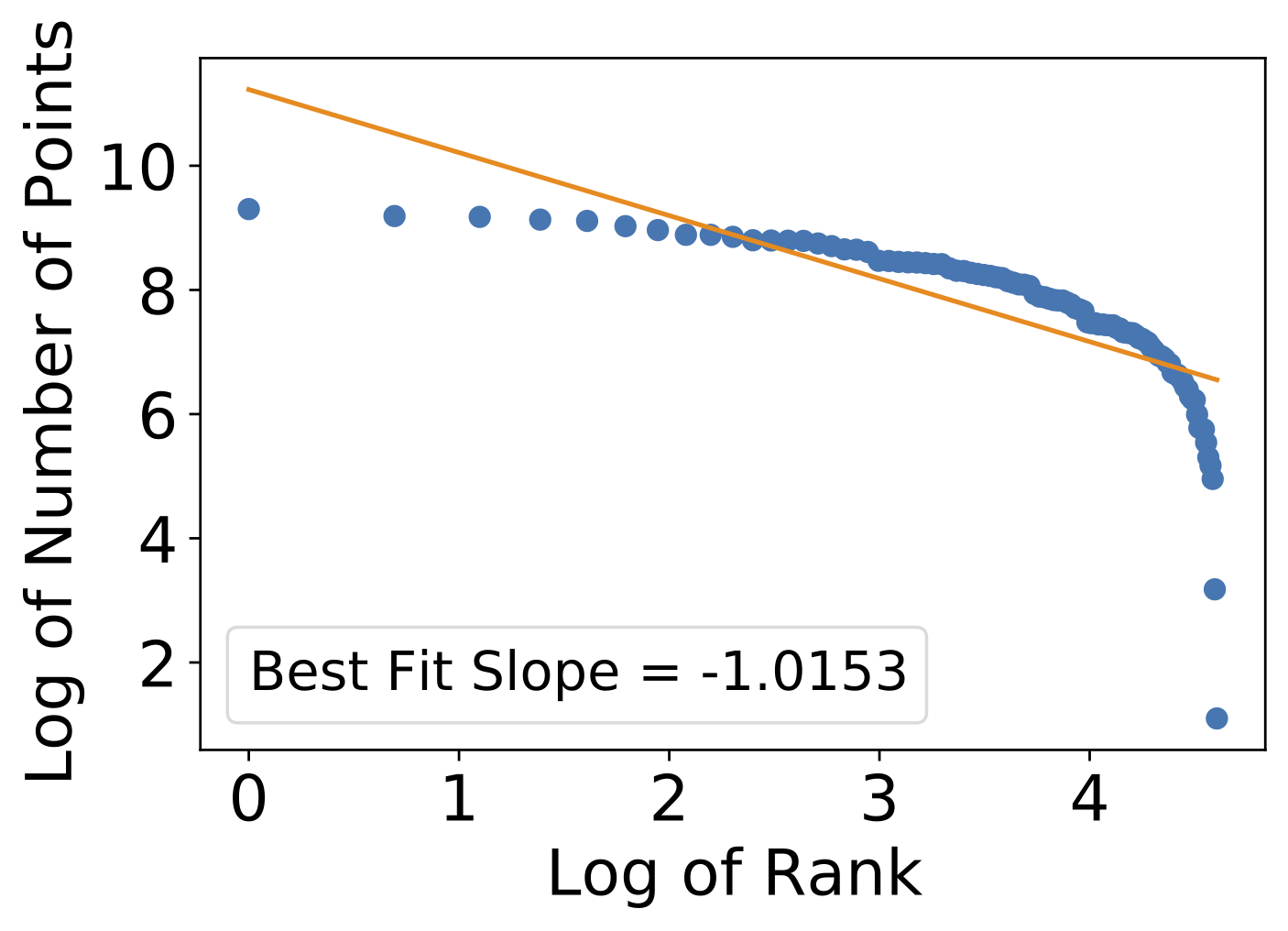}
\label{count_rank_gutenberg}}
\subfloat[RACE]{\includegraphics[width=1.25in]{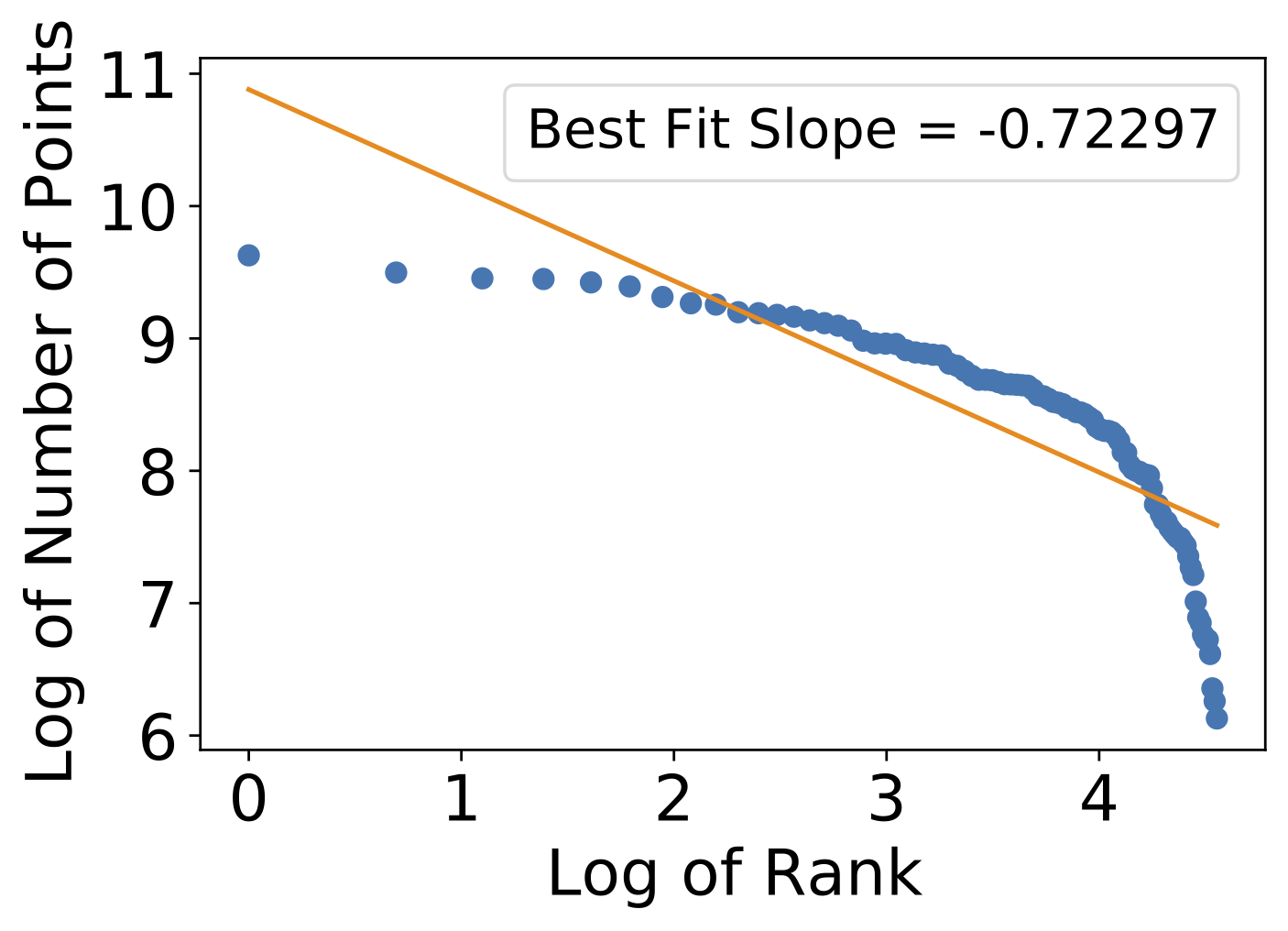}
\label{count_rank_race}}
\hfil
\vspace{-0.1in}
\caption{\small{log-log plots of cluster size vs rank, with line of best fit. Gatsby, Brown, and Gutenberg all have bets fit slopes close to -1 as in Zipf's law. The Reuters and RACE datasets are generally close but fit less well, as expected from their similar result in the categorical space.}}
\label{fig:count-rank}
\end{figure*}

After generating the clusters, we generate a log-log plot of rank of cluster size vs. cluster size as well as a line of best fit generated 
(Fig. \ref{fig:count-rank}). The slope of this fit estimates $-\alpha$ from Eq.~\ref{eq:zipf-mandelbrot}; a slope close to -1 denotes a Zipfian distribution. We also show a log-log plot of rank vs count to show how well the dataset fits the original formulation of Zipf's law (Fig. \ref{fig:original_zipfs}). 

Most datasets exhibit a certain level of heavy tails (Fig.~\ref{fig:original_zipfs}). Gatsby is a very good fit for Zipf's law, Brown is next best, and Reuters, Gutenberg, and RACE are less close. We have observe similar results in Fig. \ref{fig:count-rank}, with Gatsby and Brown showing results quite close to those expected by Zipf's law and Reuters and RACE showing more divergence. Gutenberg is more of an outlier here in that in the categorical space it fits Zipf's law about as well as Reuters and RACE but in the latent space is closer to Gatsby and Brown. This may be because the Gutenberg dataset has older text across a variety of time periods whereas the other corpora include documents created around the same time. \\ 



\subsection{Similarity Search}
\subsubsection{Baselines}
To test our set cover approach, we compare it to a number of baselines: an average of word embeddings in a sentence (AVG), the word mover's distance (WMD), the Jaccard similarity, and the Levenshtein edit distance (LD). AVG takes an unweighted average of both the query sentence and the sentences in the database and returns the $t$ sentences from the database with the lowest Euclidean distance. For the rest of the baselines, we return the $t$ sentences with the best score. If the query sentence is returned as a suggested sentence, we return the second best through $t + 1$ best results. We remove stopwords for all approaches except LD.

\subsubsection{Qualitative Results via Questionnaire}
To evaluate the efficacy of the various similarity search methods, we use the Gutenberg dataset described in Section \ref{entropy_estimation_experiment} broken up by reading level. Given a query sentence, we compare it against each sentence in our dataset and return the best $t = 5$ found according to our metric. To evaluate each approach, we create a questionnaire application that can be run from the command line and present it to users.

For the questionnaire, we first generate 100 sample queries, selected at random from the dataset at the appropriate reading level. Sentences that were too long or contained specific keywords were removed. This process was repeated for each dataset until the required number of sentences was selected: 33 each for ``pre-high school'' and ``high school'' and 34 for ``college''. Then, each algorithm was run against the query and the results were checked for readability. We also removed sentences longer than 15 tokens from our database; these longer sentences were unlikely to be helpful as they often amounted to an entire paragraph or came from ambiguous sentence breaks in poetry. 
During the questionnaire, some users were presented with a selection of five sentences, one from each baseline and one from the set cover, and asked to select the three they would find most useful while also considering variety (the choose-3 questionnaire). Other users were presented with the top five sentences from the set cover algorithm and one of the baselines (the head-to-head questionnaire). These users were split into two groups: one group chose the algorithm they thought would be most helpful in their writing, and the other group chose the algorithm that did the best job of providing variety in relevant sentences. Users were asked to answer as many questions as they could and given the option to quit after each answer. 

Participants in the experiment were students from a programming course at a large public research university (the questionnaire required basic knowledge of how to execute a Python script). Students received extra credit for participating. Altogether 107 students responded, 58 for the choose-3 survey and 49 for the head-to-head survey.
Students in the head-to-head survey answered on average 59 questions compared to an average of 52 in the choose-3 survey.

For the choose-3 survey, we found all algorithms to be selected about as frequently they would be by chance (60\%): set cover was chosen 57\% of the time, AVG 63\%, WMD 61\%, Jaccard 58\%, and LD 61\%. For the head-to-head survey, the set cover approach was preferred about 44\% of the time across all baselines and was deemed to have a better variety of related sentences 47\% of the time. The results suggest that sentences selected by our algorithm were routinely chosen, and have comparable performance to other baselines that have proven to be effective. Therefore, our novel algorithm delivered new related sentences to the students.



In our own analysis, we find the set cover suggestions are frequently more varied than those of the other algorithms (see Tables \ref{tab:example_suggestions_1} and \ref{tab:example_suggestions_2} in Appendix \ref{app:example-suggestions}). The AVG, WMD, and Jaccard algorithms often return the same results, especially for short and generic sentences. These results are generally so close to the query that they provide little new information that could be useful to a writer whereas the set cover results are related but suggest a new context or flavor.

\subsubsection{Quantitative Results}
In addition to the qualitative results presented above, we measure sentence variety in two ways. First, for each query we count the number of suggested sentences which are unique to a particular algorithm (inter-algorithm variety). This is a measure of redundancy \textit{between} algorithms, as a high percentage of non-unique suggestions indicates that an algorithm is not contributing anything new to the user. Second, we compute the pair-wise Jaccard similarity between each suggestion by a given algorithm for a given query (intra-algorithm variety). This is a measure of redundancy \textit{within} algorithms, as a high average Jaccard similarity suggests a lack of variety among the sentences suggested.

We generate these results across two datasets. The first is the 100 queries used for the questionnaire and the second is a set of 5000 queries. This second set samples sentences randomly from each reading level as queries and runs them against the database for the same reading level. The pairwise Jaccard similarity was calculated twice, once including stopwords and once with stopwords removed.

Based on the results in Table \ref{tab:unique-responses}, we can see that the set cover algorithm clearly outperforms the AVG, WMD, and Jaccard algorithms at producing unique sentences. The LD algorithm returns a comparable number of unique sentence, although we have observed it often suggests sentences seemingly unrelated to the query. Table \ref{tab:jaccard-suggestions} shows that the set cover approach offers substantial gains in intra-algorithm variety with suggestions less than half the overlap of WMD, Jaccard, and LD when stopwords are included and less than 3\% of the overlap of Jaccard when stopwords are removed. Taken together, it is clear that the set cover algorithm outperforms existing similar sentence methods on recommendation variety.

\begin{table}[t]
\renewcommand{\arraystretch}{1}
\begin{center}
\small
\caption{{\small Percent of unique suggested sentences. For 5000 queries, our approach suggested the highest number of unique sentences, more than 1.5 times the Jaccard baseline.}}
\vspace{-1mm}
\begin{tabular}{c||cccccc}
\hline
\hline
Queries & Set Cover & AVG & WMD & Jaccard & LD\\ 
\hline
100 & 89.60 & 72.00 & 42.20 & 56.40 & \textbf{93.20}\\
5000 & \textbf{89.31} & 69.89 & 48.47 & 57.07 & 88.37\\
\end{tabular}
\label{tab:unique-responses}
\vspace{-2mm}
\end{center}
\end{table}

\begin{table}[t]
\renewcommand{\arraystretch}{1}
\begin{center}
\small
\caption{\small Average pairwise Jaccard similarity among suggested results. For 5000 queries, with stopwords (Rm Stop = No) our approach contains one third as many redundant words as Jaccard and less than 3\% of Jaccard without stopwords (Rm Stop = Yes).}
\vspace{-1mm}
\begin{tabular}{cc||ccccc}
\hline
\hline
Queries & Rm Stop & Set Cover & AVG & WMD & Jaccard & LD\\ 
\hline
100 & Yes & \textbf{0.0676} & 0.1051 & 0.1437 & 0.1735 & 0.1376\\
5000 & Yes & \textbf{0.0676} & 0.1246 & 0.1728 & 0.2023 & 0.1593\\
100 & No & \textbf{0.0057} & 0.0704 & 0.1541 & 0.2648 & 0.0272\\
5000 & No & \textbf{0.0083} & 0.1034 & 0.1967 & 0.3062 & 0.0574\\
\end{tabular}
\label{tab:jaccard-suggestions}
\vspace{-2mm}
\end{center}
\end{table}

%% file: data.tex
\subsection{Datasets}
We selected five datasets for our experiments: Gatsby, Reuters, Brown, Gutenberg, and RACE. Gatsby is the text of \textit{The Great Gatsby}. This dataset was chosen to see how the results apply to a relatively small body of text written by a single author in a single style about a single topic. Reuters is a collection of 10,788 news articles on 90 different topics, all related to business from \texttt{nltk}~\cite{nltk}. Due to the short, factual nature of the newswire sentences, it is an interesting point of comparison against the other datasets which focus on more creative writing. Brown is also from \texttt{nltk}, and consists of text from 500 sources ranging from news to romance, reviews to science-fiction \cite{brown_corpus}. This corpus was primarily included as it is a standard in many textual analysis and NLP applications, including many studies of Zipf's Law \cite{zipf_brown, zipf_brown_2}. Gutenberg is a collection of 79 books from Project Gutenberg.\footnote{\url{https://www.gutenberg.org/}} As Gutenberg only contains works which are outside of copyright, the dataset is mostly older (pre-1940). All books were chosen are those likely to be common in curricula in United States education from middle school through college and should be familiar to most readers. Finally, the RACE Dataset \cite{RACE}, obtained from the project website,\footnote{\url{http://www.cs.cmu.edu/~glai1/data/race/}} is a collection of English text used to teach Chinese students in middle and high schools. We use the article text, which varies from simple narrative sentences to short biographies and articles. This dataset is unique due to its size, variety, and the fact that it was created by bi-lingual authors for the purpose of teaching English as a second language.

Words and sentences were tokenized using the standard \texttt{nltk} tokenizers and embedded using fastText \cite{fasttext}. 
For basic descriptive statistics of these datasets, refer to Table \ref{tab:fasttext_description} in Appendix \ref{app:dataset-details}.


%% file: part_of_speech.tex
We used the standard $\texttt{nltk}$ part of speech tagger \cite{nltk} and reduced the tags into six categories: noun, verb, adjective, adverb, pronoun, and other. This removes some of the divisions not present in \cite{zipf_latent} such as singular and plural nouns. It also combines a number of tags with few unique words, as the analysis on embeddings does not capture frequency.

First, we look at the composition of the tags in each of the five datasets. We see that, as expected, in each dataset the ``other" category which includes articles has more appearances than any other part of speech. See Appendix \ref{app:part-of-speech} for a full break-down.

Next, we investigate whether words are embedded in the latent space next to words of the same part of speech. This is parameterized by $k$, the number of neighbors to consider in the calculation. The results are reported in Table \ref{pos_nbrs}. As we can see, in all of the datasets and for all tested values of $k$, on average at least half of a word's neighbors in the latent space are of the same part of speech. When looking at only the three closest words, on average at least 2/3 of the neighbors are of the same part of speech. 


Fig.~\ref{fig:tsne} visualizes the relationship between values in the latent space and part of speech, using t-distributed stochastic neighbor embedding (t-SNE) \cite{t-sne} to plot the embeddings in two dimensions. We also performed Principal Component Analysis (PCA) \cite{pca} and multidimensional scaling (MDS) \cite{mds}, but found t-SNE to be the most informative.
As we can see, points generally seem to be grouped by part of speech, although there are many exceptions. This supports the numerical analysis: words in the latent are most frequently associated with words of the same part of speech but not exclusively so. It is worth recalling that because there is only one embedding per word, the embedding must capture all parts of speech associated with this word simultaneously. This nuance is not captured in the plot as the part of speech tagger assigns only a single part of speech for each word and it is not clear how this is represented in the original latent space or the latent space after performing t-SNE. It may be that neighbors with different parts of speech come from the same root word, e.g., slows (verb), slowly (adverb), slow (adjective).


%% file: conclusion.tex
\section{Conclusion and Future Work}
In this paper, we examined how we can revitalize low compute-cost word-level NLP toolkits (originally designed for bag-of-words representations) under embedding models. We answered three important questions: \emph{(i)} design of a non-parametric KL divergence estimation algorithms based on $k$-NN methods, \emph{(ii)} analysis of power law behaviors of the embedded points using part-of-speech and clustering tools, and \emph{(iii)} design of a combinatorial algorithm based on set covers for identifying similar/related sentences. Extensive experiments showed that our computationally efficient algorithms are effective for various NLP tasks. 

Important future works include generalizing our results to other languages, optimizing our algorithms towards a specific architecture (e.g., GPU), and building full-fledged data mining applications (e.g., author attribution, creative writing assistance) by integrating our techniques with existing modeling tools. 

%% file: appendix.tex
\appendix
\section{Dataset Details}
\label{app:dataset-details}

Table \ref{tab:fasttext_description} presents additional information on the datasets. In this table, unembedded tokens refers to the number of unique tokens found in the text for which there was no fastText embedding available, even after correcting for some common issues like errant periods. Unique sentences removes sentences for which no word could be embedded (fewer than ten in each dataset) and repeated sentences.
\begin{table}[!t]
\renewcommand{\arraystretch}{1.3}
\begin{center}
\small
\caption{\small Descriptive statistics for FastText embeddings.}
\vspace{-2mm}
\begin{tabular}{p{1.2cm}||p{.9cm}p{1cm}p{1.25cm}p{1.2cm}}
\hline
\hline
Dataset & Total Tokens & Unique Tokens & Unembedded Tokens & Unique Sentences\\ 
\hline
Gastby & 58,339 & 6,291 & 70 & 3,417 \\
Reuters & 1,712,123 & 34,712 & 5,311 & 51,555 \\
Brown & 1,15,9118 & 49,642 & 2,183 & 56,416 \\
Gutenberg & 7,454,306 & 76,466 & 23,544 & 306,741\\
RACE & 9,031,066 & 89,076 & 16,854 & 492,921 \\
\end{tabular}
\label{tab:fasttext_description}
\vspace{1mm}
\end{center}
\end{table}

\section{Part of Speech}
\label{app:part-of-speech}
\begin{figure}[!t]
\centering
\caption{\small{t-SNE plots of unique words, colored by part of speech, for the two smaller datasets. See Figure \ref{fig:tsne} for a legend. The same results hold: words are frequently close to words with the same part of speech. This suggests the latent variable which may account for Zipf's law in the categorical case is also observable in the word embeddings, even for smaller corpora.}}
\subfloat[Gatsby]{\includegraphics[width=2.1in]{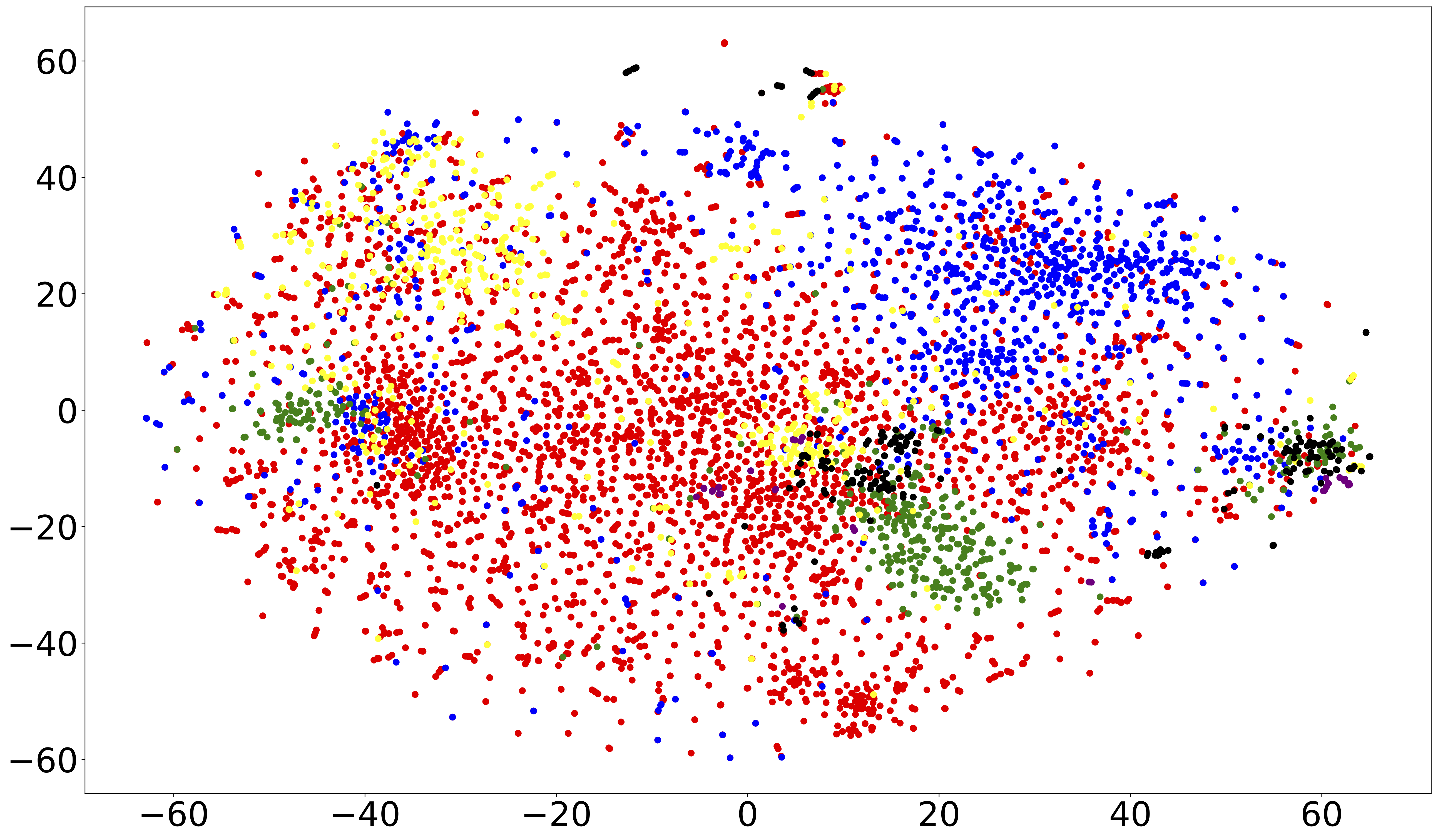}
\label{tsne_gatsby}}
\hfil
\subfloat[Reuters]{\includegraphics[width=2.1in]{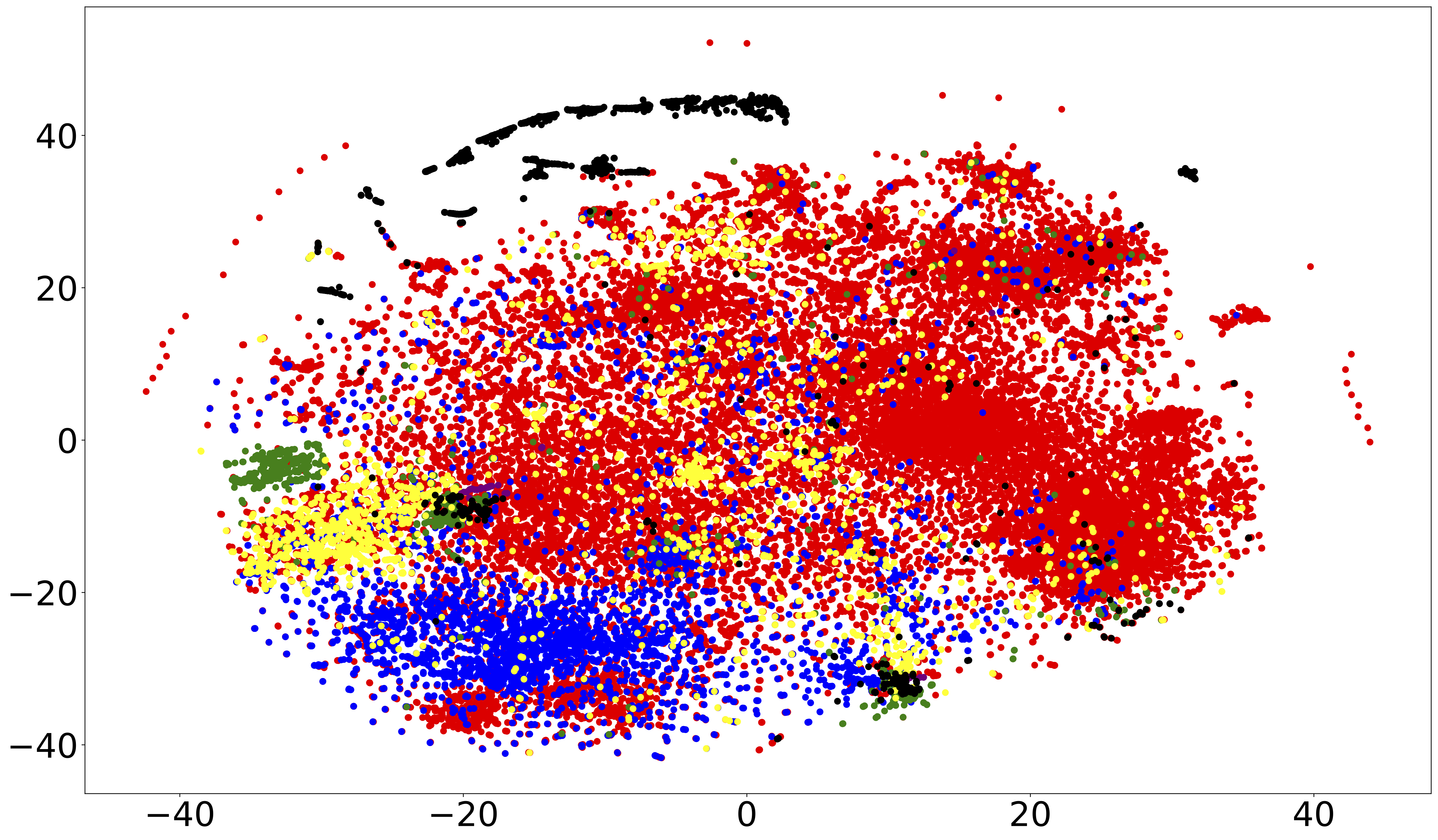}
\label{tsne_reuters}}
\label{fig:tsne_small}
\end{figure}

This appendix presents the part of speech statistics for each dataset in Table \ref{pos_description} and the t-SNE plots for the Gatsby and Reuters datasets in Fig. \ref{fig:tsne_small}.

\begin{table*}[!t]
\renewcommand{\arraystretch}{1.1}
\begin{center}
\small
\caption{\small Part of speech statistics. These are fairly consistent despite the difference in size and content among corpora and align well with the conditions for the latent variable explanation of Zipf's law.}
\vspace{-1mm}
\begin{tabular}{c||cccccc}
\hline
\hline
&\multicolumn{6}{c}{Total Tokens (Percent of Total)}\\
Dataset & Nouns & Verbs & Advs & Adjs & Pronouns & Other\\ 
\hline
Gatsby & 14,679 (25.16\%) & 7,357 (12.61\%) & 3,548 (6.08\%) & 1,996 (3.42\%) & 6,333 (10.86\%) & 24,426 (41.87\%)\\
Reuters & 664,378 (38.80\%) & 144,512 (8.44\%) & 42,382 (2.47\%) & 71,368 (4.17\%) & 32,164 (1.88\%) & 757,621 (44.24\%)\\
Brown & 345,489 (29.59\%) & 129,046 (11.05\%) & 59,770 (5.12\%) & 60,021 (5.14\%) & 63,611 (5.45\%) & 509,671 (43.65\%)\\
Gutenberg & 2,048,937 (27.49\%) & 853,348 (11.45\%) & 452,293 (6.07\%) & 270,644 (3.63\%) & 687,315 (9.22\%) & 3,141,769 (42.15\%)\\
RACE & 2,709,018 (30.01\%) & 1,043,036 (11.55\%) & 501,481 (5.55\%) & 462,794 (5.13\%) & 637,996 (7.07\%) & 3,673,762 (40.69\%)\\
\hline
\hline
& \multicolumn{6}{c}{Unique Tokens (Percent of Unique)}\\
Dataset & Nouns & Verbs & Advs & Adjs & Pronouns & Other\\
\hline
Gatsby & 3,858 (61.33\%) & 1,210 (19.23\%) & 465 (7.39\%) & 509 (8.09\%) & 37 (0.59\%) & 212 (3.37\%)\\
Reuters  & 26,435 (76.19\%) & 3,734 (10.76\%) & 809 (2.33\%) & 1,780 (5.13\%) & 40 (0.12\%) & 1,898 (5.47\%)\\
Brown  & 35,161 (70.84\%) & 6,715 (13.53\%) & 1,864 (3.76\%) & 4,327 (8.72\%) & 40 (0.08\%) & 1,530 (3.08\%)\\
Gutenberg & 56,563 (73.97\%) & 9,974 (13.04\%) & 2,804 (3.67\%) & 5,631 (7.36\%) & 52 (0.07\%) & 1,442 (1.89\%)\\
RACE  & 66,089 (74.19\%) & 8,400 (9.43\%) & 2,065 (2.32\%) & 7,787 (8.74\%) & 52 (0.06\%) & 4,682 (5.26\%)\\
\vspace{-3mm}
\end{tabular}
\label{pos_description}
\end{center}
\end{table*}
\vspace{-2mm}

\section{Zipf's Law Robustness to \textit{k}}
\label{app:zipf-changes-k}

\begin{figure}[!t]
    \centering
    \includegraphics[width=2.5in]{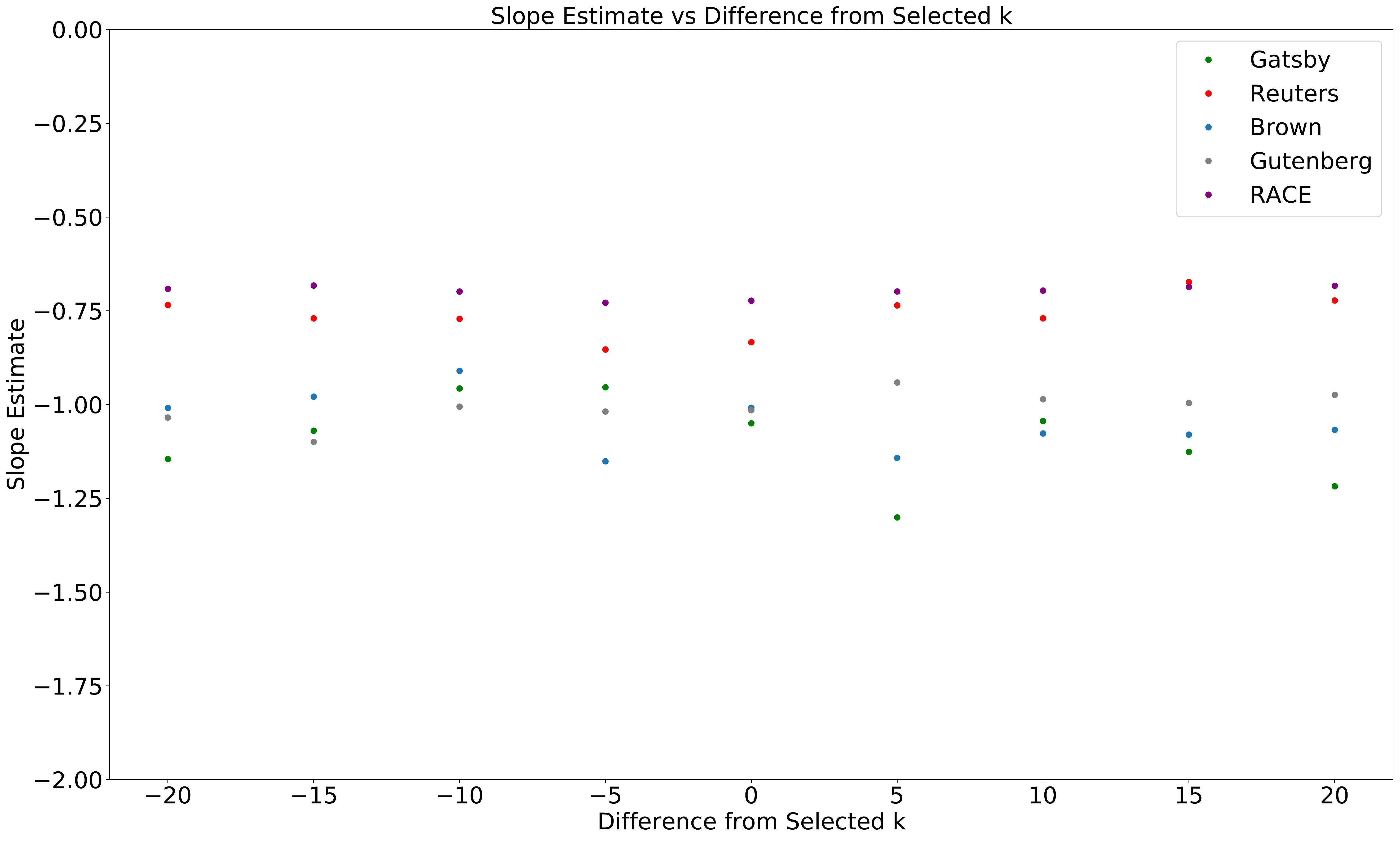}
    \caption{\small Changes in best fit slope relative to selected $k$. Even for quite significant changes in $k$, the best fit slopes do not drastically change.}
    \label{fig:k_with_slope}
\end{figure}


In Figure~\ref{fig:k_with_slope}, we plot the slope of the best fit line for values of $k$ ranging from $k' - 20$ to $k' + 20$ in multiples of 5 where $k'$ is the value fine-tuned for that dataset. In general, the slopes do not change enough to alter our conclusions about the applicability of a Zipfian distribution. The Gatsby dataset appears to have the most variance, but this is most likely because it is the shortest and has a small $k'$ (35) to begin with.

\section{Example Suggested Sentences}
\label{app:example-suggestions}
A few examples of results from our algorithm and the baselines can be found in Tables \ref{tab:example_suggestions_1} and \ref{tab:example_suggestions_2}.

\begin{table}[!t]
\footnotesize
\caption{\small Example Suggested Sentences 1. Note the repetition and blandness of the baselines compared to the set cover.}
\vspace{-2mm}
\label{tab:example_suggestions_1}
\begin{tabular}{  p{1cm}  p{3cm} p{3cm}}
\toprule
\textbf{Model}      
& ``Come along," she said. & ``Let me look at you."
\\\midrule
Set Cover 
&
\begin{enumerate}[leftmargin=*]
    \vspace{-1.75mm}
    \item Look at this, says he.
    \item And then she told him to say papa.
    \item Coming abreast of the calèche he ran beside it.
    \item No-one would have thought the end was so near.
    \item What was he saying?
\end{enumerate} &
\begin{enumerate}[leftmargin=*]
    \vspace{-1.75mm}
    \item Go on; let me see.
    \item Let us say so.
    \item You look like communards.
    \item Come, go along!
    \item Hélène stooped forward to make room, and looked round with a smile.
\end{enumerate}\\

Embedding Average      
& \begin{enumerate}[leftmargin=*]
    \vspace{-1.75mm}
    \item ``Come along, our Matvévna!" he said to himself.
    \item ``Come in, come in," he said."
    \item Come along.
    \item Come along!
    \item Come along now.
\end{enumerate}
& \begin{enumerate}[leftmargin=*]
    \vspace{-1.75mm}
    \item Let me have a look at it....
    \item Let me look.
    \item Let go of me.
    \item Let it go.
    \item —Tiptop... Let me see.
\end{enumerate}\\
Word Mover's Distance        
& \begin{enumerate}[leftmargin=*]
    \vspace{-1.75mm}
    \item ``Come along, our Matvévna!" he said to himself.
    \item ``Come in here," she said."
    \item Come, go along!
    \item Come, come along!
    \item ``Come along then!"
\end{enumerate}
& \begin{enumerate}[leftmargin=*]
    \vspace{-1.75mm}
      \item Let me have a look at it....
      \item Let me look.
      \item Let it go.
      \item Let go of me.
      \item Let me see; where was I?
\end{enumerate}\\
Jaccard &\begin{enumerate}[leftmargin=*]
    \vspace{-1.75mm}
    \item``Come along, our Matvévna!" he said to himself.
    \item ``Come along then!"
    \item Come along!
    \item Come along.
    \item Come along now.
\end{enumerate}
&\begin{enumerate}[leftmargin=*]
    \vspace{-1.75mm}
      \item Let me have a look at it....
      \item Let me look.
      \item ``Let him look for me," thought she.
      \item Is that how you look at it?
      \item ``Let them!"
\end{enumerate}\\
Levenshtein Distance &
\begin{enumerate}[leftmargin=*]
    \vspace{-1.75mm}
    \item —Come, he said.
    \item ``Come along then!"
    \item Come along.
    \item Come along now.
    \item Good again, he said.
\end{enumerate}
 & \begin{enumerate}[leftmargin=*]
    \vspace{-1.75mm}
      \item Let me look.
      \item Sees me looking.
      \item Let me hear.
      \item ``Let me be!...",
      \item Dress they look at.
\end{enumerate}\\
\bottomrule
\end{tabular}
\end{table}

\begin{table}[!t]
\footnotesize
\caption{\small Example Suggested Sentences 2. The Levenshtein distance results often appear unrelated to the query. For the first query, AVG and WMD share a sentence, as do WMD and Jaccard. For the second, the set cover suggests related but diverse sentences one could imagine using to continue a story whereas baselines tend to focus on the words "well" and "mind." AVG returns some quality sentences in each, e.g. suggestion 4 for each query, but also offers some seemingly unrelated such as 5 for query one and 3 for query two.}
\vspace{-1mm}
\label{tab:example_suggestions_2}
\begin{tabular}{  p{1cm}  p{3cm} p{3cm}}
\toprule
\textbf{Model}      
& People laughed particularly at his passion for psychology. & A real machine; well, I don't mind serving a machine.
\\\midrule
Set Cover 
&
\begin{enumerate}[leftmargin=*]
    \vspace{-1.75mm}
    \item cried Orlando still laughing.
    \item Psychology lures even most serious people into romancing, and quite unconsciously.
    \item ``Ha-ha-ha!" laughed Pierre.
    \item It's passion, not love.
    \item He gave himself up to every new idea with passionate enthusiasm.
\end{enumerate} &
\begin{enumerate}[leftmargin=*]
    \vspace{-1.75mm}
    \item Laurie, though decidedly amazed, behaved with great presence of mind.
    \item ``Very well indeed!"
    \item She thought it served Charley right, too.
    \item Every man was tearing himself loose, even Matthewson.
    \item ``I, also, am of your way of thinking."
\end{enumerate}\\

Embedding Average      
& \begin{enumerate}[leftmargin=*]
    \vspace{-1.75mm}
    \item ``He brought in too much psychology," said another voice.
    \item ``Yes, yes!"
    \item People said that his poetry was ``so beautiful."
    \item ``The only subjects I respect are mathematics and natural science," said Kolya.
    \item Everyone spoke loudly of the field marshal's great weakness and failing health.
\end{enumerate}
& \begin{enumerate}[leftmargin=*]
    \vspace{-1.75mm}
    \item he said.
    \item ``You haven't seen my private expense book yet."
    \item ``It's clear that you don't know how large the sum is."
    \item ``Now, isn't it a dreadful state of things?"
    \item ``It's very pretty -- new thing, isn't it?"
\end{enumerate}\\
Word Mover's Distance        
& \begin{enumerate}[leftmargin=*]
    \vspace{-1.75mm}
    \item People laughed.
    \item People said that his poetry was ``so beautiful."
    \item People in law perhaps.
    \item People looked at him with hatred.
    \item O, I never laughed so many!
\end{enumerate}
& \begin{enumerate}[leftmargin=*]
    \vspace{-1.75mm}
    \item ``Didn't I do well?"
    \item ``I don't mind it, and he needn't know."
    \item ``Oh, I don't mind him a bit," said Dorothy.
    \item ``Well, well, he made a mistake, didn't he?"
    \item Didn't that do as well as a regular prayer?
\end{enumerate}\\
Jaccard &\begin{enumerate}[leftmargin=*]
    \vspace{-1.75mm}
    \item People laughed.
    \item How she laughed!
    \item She laughed.
    \item All laughed.
    \item All we laughed.
\end{enumerate}
&\begin{enumerate}[leftmargin=*]
    \vspace{-1.75mm}
    \item ``Don't mind me."
    \item ``Didn't I do well?"
    \item ``I don't mind it, and he needn't know."
    \item ``Well, I will, if Mother doesn't mind."
    \item Didn't that do as well as a regular prayer?
\end{enumerate}\\
Levenshtein Distance &
\begin{enumerate}[leftmargin=*]
    \vspace{-1.75mm}
    \item She was particularly fond of Smurov.
    \item He passed Saint Joseph's National school.
    \item People talk about you a bit: forget you.
    \item Gallaher, that was a pressman for you.
    \item She blushed continually and was irritable.
\end{enumerate}
 & \begin{enumerate}[leftmargin=*]
    \vspace{-1.75mm}
    \item ``The machine was standing on a sloping beach."
    \item I declared I wouldn't and got mad.
    \item All things were thawing, bending, snapping.
    \item I see no reason for hindering them.
    \item Grey Beaver was a god, and strong.
\end{enumerate}\\
\bottomrule
\end{tabular}
\end{table}

\clearpage